\documentclass{article}

\PassOptionsToPackage{numbers, sort&compress}{natbib}
\usepackage[dblblindworkshop, final]{neurips_2025}
\workshoptitle{Space in Vision, Language, Embodied AI}

\usepackage[utf8]{inputenc} 
\usepackage[T1]{fontenc}    
\usepackage{hyperref}       
\usepackage{url}            
\usepackage{booktabs}       
\usepackage{amsfonts}       
\usepackage{nicefrac}       
\usepackage{microtype}      
\usepackage{xcolor}         
\usepackage{subfigure}
\usepackage{graphicx}
\usepackage{array}
\usepackage{booktabs}
\usepackage{colortbl}
\usepackage[table]{xcolor}
\usepackage{geometry}
\usepackage{listings}
\usepackage{xcolor}

\title{See it. Say it. Sorted: Agentic System for Compositional Diagram Generation}

\author{
  Hantao Zhang\\
  Yale University\\
  \texttt{hantao.zhang@yale.edu} \\
  \And
  Jingyang Liu\\
  University of Edinburgh \\
  \texttt{jingyang\_\_liu@outlook.com} \\
  \And
  Ed Li\\
  Yale University\\
  \texttt{ed.li@yale.edu} \\
}

\begin{document}

\maketitle

\begin{figure}[h!]
    \centering
    \includegraphics[width=0.85\textwidth]{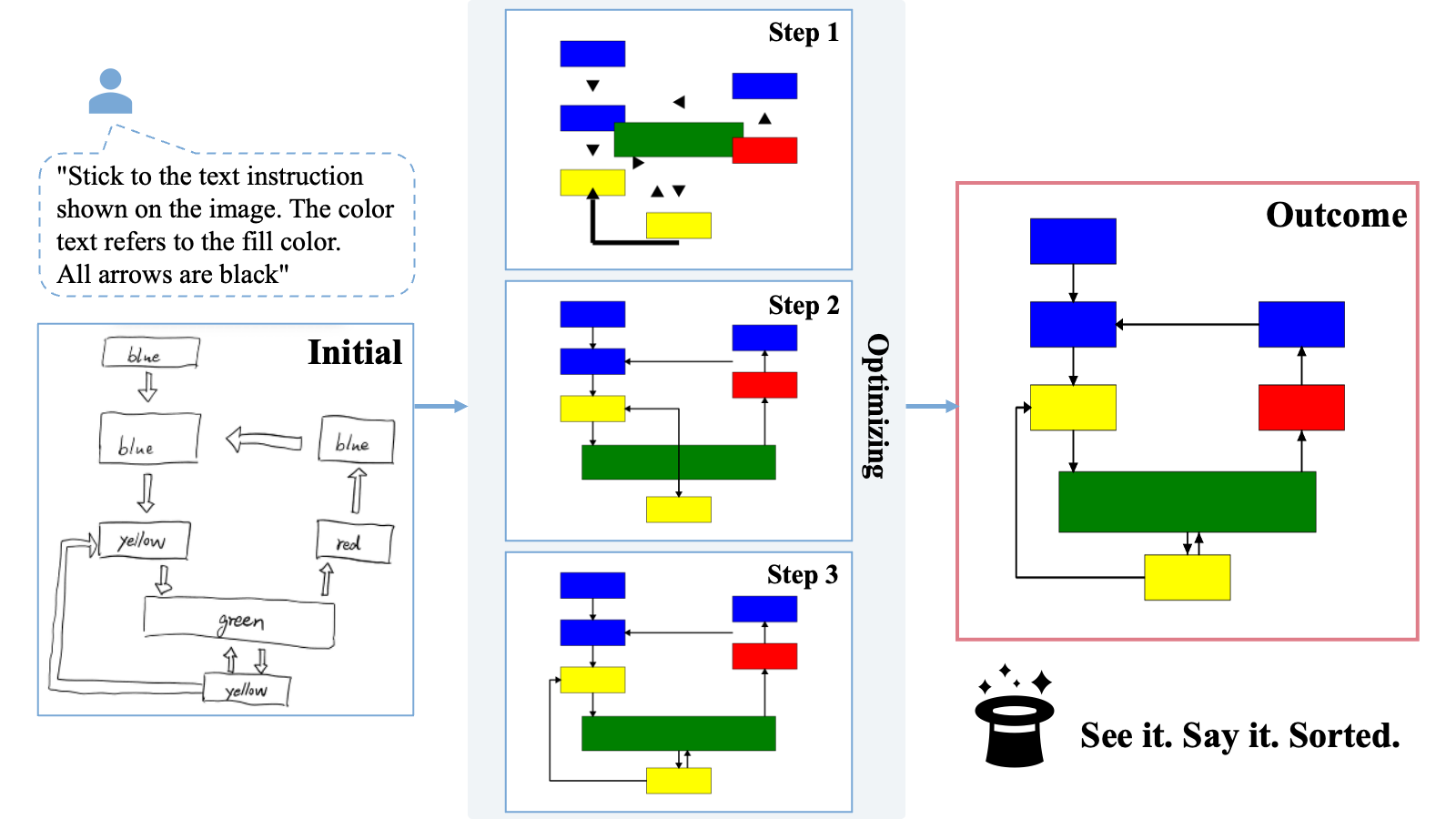}
    \caption{Given a sketch of a flow chart, our agent faithfully reconstructs the intended structure while strictly following the accompanying text instructions. The system operates in an iterative Critic–Candidates–Judge loop: a VLM critiques discrepancies between the sketch and the current diagram, multiple LLMs propose diverse SVG modifications, and a Judge VLM selects the best candidate. This training-free framework enables accurate, controllable, and editable diagram generation, moving beyond pixel-level synthesis toward structured programmatic outputs.}
    \label{fig:optimization_process}
\end{figure}

\begin{abstract}
  We study \emph{sketch-to-diagram} generation: converting rough hand sketches into precise, compositional diagrams. Diffusion models excel at photorealism but struggle with the spatial precision, alignment, and symbolic structure required for flowcharts. We introduce \emph{See it. Say it. Sorted.}, a \emph{training-free} agentic system that couples a Vision–Language Model (VLM) with Large Language Models (LLMs) to produce editable Scalable Vector Graphics (SVG) programs. The system runs an iterative loop in which a \emph{Critic VLM} proposes a small set of qualitative, relational edits; multiple \emph{candidate LLMs} synthesize SVG updates with diverse strategies (conservative$\rightarrow$aggressive, alternative, focused); and a \emph{Judge VLM} selects the best candidate, ensuring stable improvement. This design prioritizes qualitative reasoning over brittle numerical estimates, preserves global constraints (e.g., alignment, connectivity), and naturally supports human-in-the-loop corrections. On 10 sketches derived from flowcharts in published papers, our method more faithfully reconstructs layout and structure than two frontier closed-source image generation LLMs (GPT-5 and Gemini-2.5-Pro), accurately composing primitives (e.g., multi-headed arrows) without inserting unwanted text. Because outputs are programmatic SVGs, the approach is readily extensible to presentation tools (e.g., PowerPoint) via APIs and can be specialized with improved prompts and task-specific tools. The codebase is open-sourced at \href{https://github.com/hantaoZhangrichard/see_it_say_it_sorted.git}{\texttt{https://github.com/hantaoZhangrichard/see\_it\_say\_it\_sorted.git}.}
\end{abstract}

\section{Introduction}
Sketch-to-diagram generation transforms rough hand-drawn sketches into precise, compositional diagrams or flowcharts, akin to those made in tools like PowerPoint. This has broad applications in software engineering, education, and design, particularly in collaborative settings requiring rapid ideation and clear communication.

Diffusion-based generative models achieve high-fidelity image synthesis from text and sketches \cite{zhang_2023_adding, kawar_imagic, mou_2023_t2iadapter}, but struggle with the compositional reasoning and spatial precision needed for structured diagrams. Their pixel-based diffusion process makes it difficult to maintain distinct shapes, lines, and alignments \cite{koley_2023_picture, zhang_2024_sketchguided}, and they are not easily adaptable to varying diagram sizes or fine-grained user instructions.

Meanwhile, Vision–Language Models (VLMs) and Large Language Models (LLMs) have shown strong \emph{2D spatial reasoning} \cite{yamada_2023_evaluating, wu_2024_visualizationofthought, cai_an}. Agentic VLM+LLM systems have been applied to visual-feedback tasks like 3D graphics editing \cite{huang_2024_blenderalchemy, gu_2025_blendergym}, part assembly \cite{yamada_2024_l3go}, and scene generation \cite{hu_2024_scenecraft}, with the VLM acting as critic and the LLM generating code. Other works train transformers to produce graphics formats like Scalable Vector Graphics (SVG) and Constructive Solid Geometry (CSG) \cite{carlier_2020_deepsvg, wu_chat2svg, kapur_2024_diffusion}, though these lack the generalization of frontier VLMs.

We introduce a \emph{training-free} VLM+LLM agentic system, \emph{See it. Say it. Sorted.}, which interprets sketches and text instructions to produce editable SVG diagrams. Compared to diffusion methods, our approach offers:
\begin{itemize}
    \item \textbf{Adaptability:} Handles varying sizes and complexities without retraining.
    \item \textbf{Human-in-the-loop:} Allows iterative refinement at any stage.
    \item \textbf{Interpretability:} Outputs object-based, modifiable SVG code.
\end{itemize}

These properties make our approach highly extensible to real-world graphics design environments, such as PowerPoint or Google Slides, by integrating with their APIs. As a proof-of-concept, our work demonstrates how an agentic system can serve as a foundation for compositional graphics design—moving beyond pixel-level synthesis toward structured, controllable, and user-centric workflows. The framework can be further specialized through refined prompt design and the integration of pre-built, task-specific tools, enabling tailored solutions for a wide range of application domains.

\section{See it. Say it. Sorted.}

\begin{figure}[h!]
    \centering
    \includegraphics[width=0.8\textwidth]{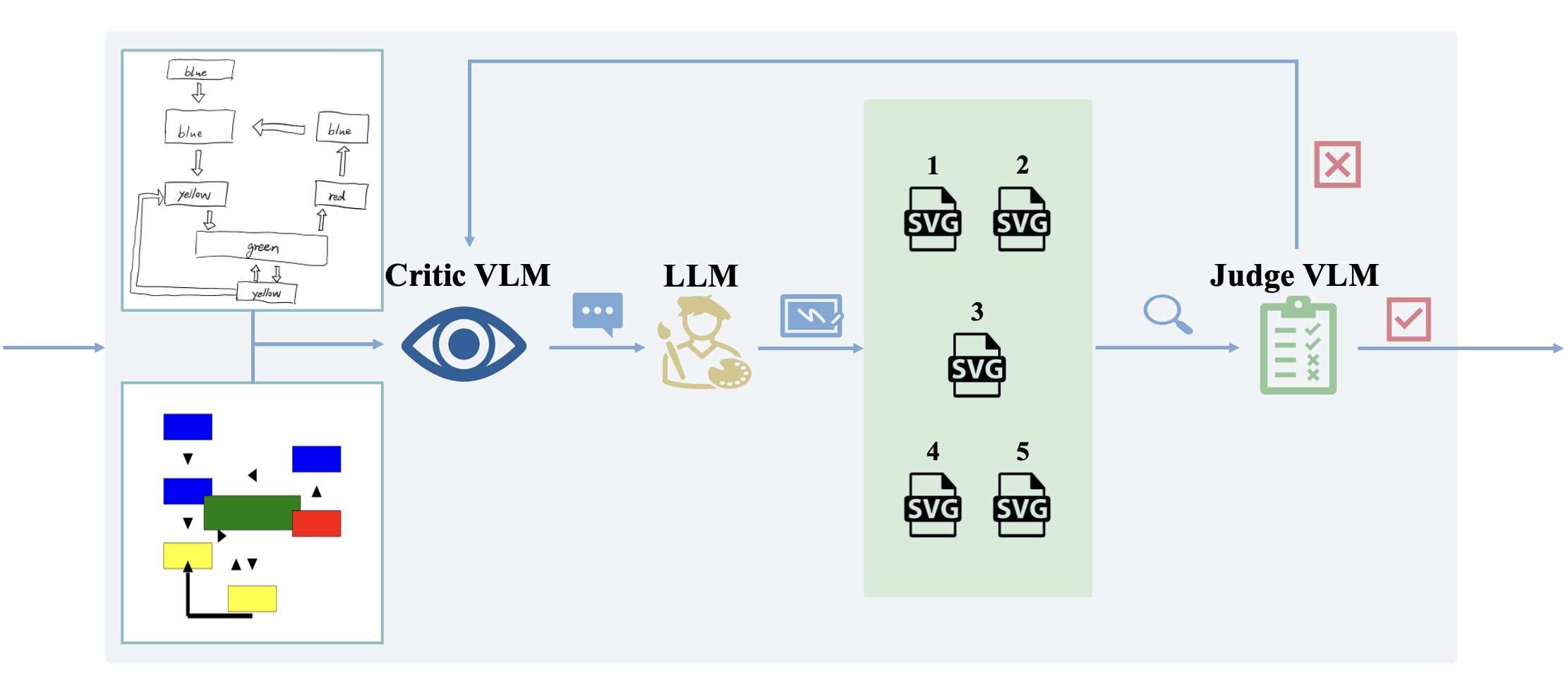}
    \caption{Pipeline of \emph{See it. Say it. Sorted.} for one optimization step: The Critic VLM compares the target sketch and the current image and identifies few small modifications, passing to LLM. Given VLM's instructions, LLM generates several candidates that balances exploration-exploitation trade-off. These candidates are rendered and evaluated together with the current image by the Judge VLM. Judge VLM decides which one best reconstruct the sketch. If the current image is chosen, then the optimization is reverted and Critic VLM receives feedback of failed modifications. If one of the candidates is chosen then proceed to next optimization step.}
    \label{fig:pipeline}
\end{figure}

We present our agentic system for sketch-to-diagram generation and evaluate its effectiveness on SVG graphics generation. We only equip our agent with minimal set of primitive shapes and color palette, details shown in Appendix \ref{svg_grammar}. Given this limitation, our agent is still able to faithfully reconstruct the sketch image.  Our results are compared against other image generation methods using 10 sketches derived from flowcharts in real published papers.

\paragraph{Initial program.}  
The \emph{Critic VLM} examines the target sketch and describes it in terms of available primitives, including approximate positions, sizes, and colors. This qualitative description is passed to an LLM, which generates an initial SVG expression as the starting point for refinement.

\paragraph{See it. Say it.}  
At each optimization step, the Critic VLM compares the target and current images, then: 1. Gives a high-level scene description; 2. Identifies 1--3 key discrepancies; 3. Suggests targeted modifications.

This prompting strategy follows two design principles:

\textit{(i) Limiting modifications per step.}  
Restricting the VLM to only a few changes per iteration helps stabilize the optimization process. Without this constraint, VLMs often produce large sets of suggestions, mixing correct and incorrect ones. Applying all changes at once can cause the generated image to oscillate between suboptimal states.

\textit{(ii) Emphasizing qualitative over quantitative descriptions.}  
Rather than requesting exact coordinates or dimensions, we instruct the VLM to express relative spatial and size relationships, referencing other primitives or the canvas as anchors (e.g., \emph{``The width should be about half the canvas; the size should be doubled; the blue rectangle should just touch the red circle on its left.''}). This choice is motivated by:
\begin{itemize}
    \item VLMs are more reliable at describing qualitative relations than at estimating precise numerical values \cite{chen_2024_spatialvlm}.
    \item The generated image evolves over time, with primitives moving, being added, or removed, making a fixed external ID mapping unreliable.
    \item Qualitative references provide richer contextual cues, enabling the LLM to identify the intended primitive more accurately.
    \item This approach gives the LLM greater flexibility in exploring the solution space, avoiding overly rigid adherence to potentially inaccurate quantitative instructions.
\end{itemize}

\paragraph{Sorted.}  
Given the VLM’s feedback, multiple LLMs generate candidate SVGs in JSON format following the grammar specified in \ref{svg_grammar} using distinct strategies (e.g., conservative, moderate, aggressive, alternative, focused) to balance exploration and exploitation. LLMs are instructed to respect the VLM’s general description to maintain global constraints such as alignment and connectivity. These candidates are then rendered into images, passing to the Judge VLM.

The \emph{Judge VLM} evaluates the candidates and current image, selecting the one that best matches the sketch. If no candidate improves the result, the step is reverted and the Critic VLM revises its suggestions. This loop of critique, diverse modification, and selection enables stable convergence toward the target diagram.

\begin{figure}[h!]
    \centering  
    \includegraphics[width=0.8\textwidth]{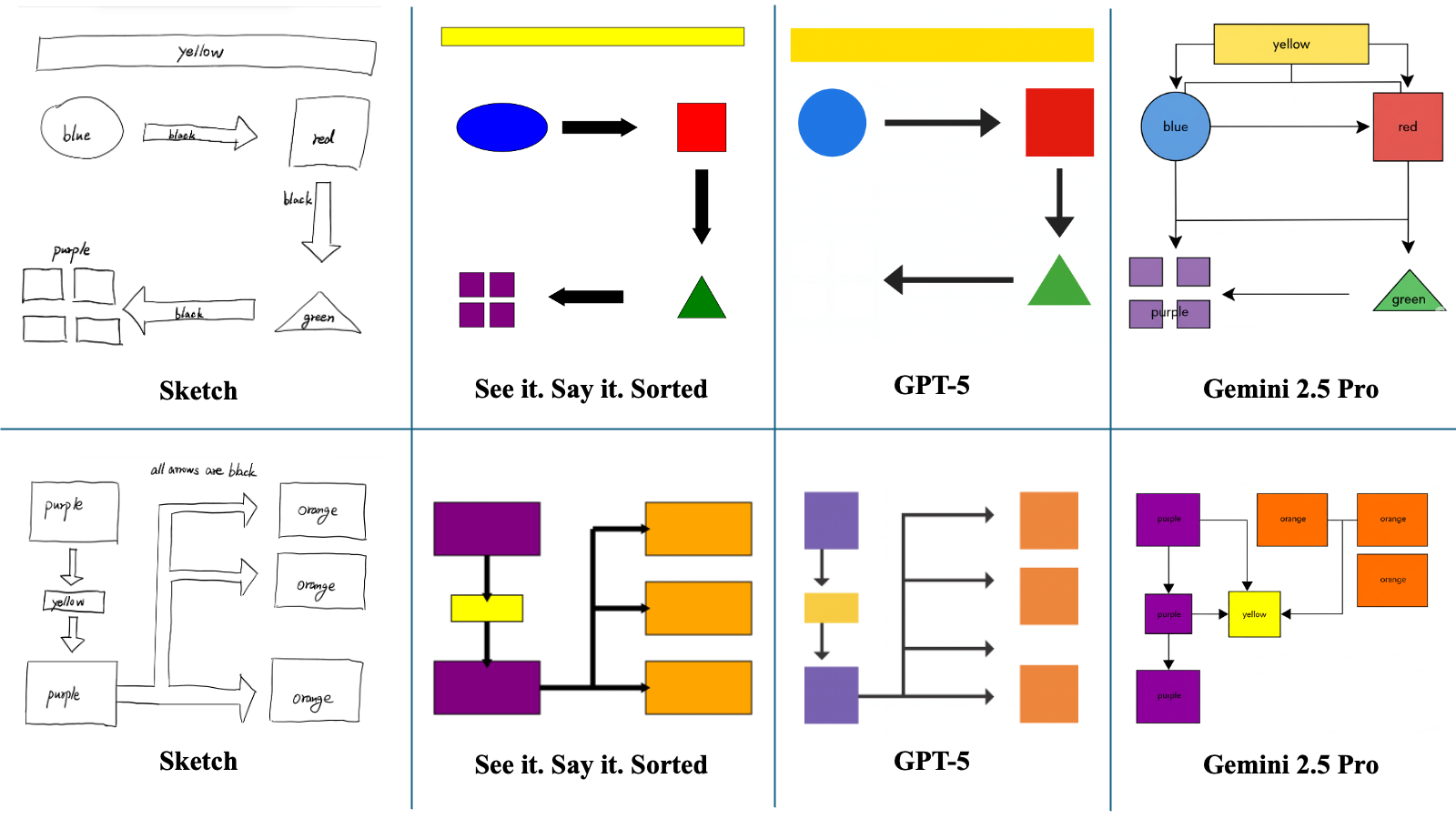}
    \caption{Our agent faithfully generates flow charts based on the sketch within 3 optimization steps, significantly outperforming GPT-5 and Gemini-2.5-Pro at preserving the structure and characteristics of the diagram. For the complete comparison for all 10 tasks, see Appendix \ref{tab:comparison_1}.}
    \label{fig:results_comparison}
\end{figure}

\section{Experiment Results}
We evaluated our agent on 10 sketch images inspired by flowcharts from real published papers. Each sketch was accompanied by a single text instruction: \emph{``Stick to the text instruction shown on the image. The color text refers to the fill color. Do not include any text in the final diagram.''}. In all experiments, we used Gemini-2.5-Pro as Critic VLM and Judge VLM and Gemini-2.5-Flash as the LLM.

Figure~\ref{fig:results_comparison} compares our system against two frontier closed-source image generation models: GPT-5 and Gemini-2.5-Pro. Our agent consistently reconstructs the structure depicted in the sketch, maintaining precise horizontal and vertical alignments between blocks. Notably, it can use fundamental primitives (e.g., rectangles, triangles) to compose arrows, including complex multi-headed arrows with correct orientations. 

In contrast, GPT-5 often misses key structural details—for example, omitting an entire cluster of purple blocks (top-row example) or producing an incorrect number of arrowheads (bottom-row example). Gemini-2.5-Pro exhibits more severe failures, frequently generating layouts that diverge from the sketch. Even when explicitly instructed that the text in the sketch only indicates fill colors and should not appear in the final diagram, Gemini repeatedly inserts text into blocks.

\section{Discussion}
We presented \emph{See it. Say it. Sorted.}, a \emph{training-free} VLM+LLM agentic system that transforms hand-drawn sketches into structured, editable SVG diagrams through an iterative Critic–Candidates–Judge loop. By emphasizing qualitative, relational feedback rather than brittle numerical estimates, the method preserves global structural constraints such as alignment and connectivity while enabling precise local refinements, outperforming frontier image LLMs in faithfully reconstructing compositional layouts. Because outputs are programmatic, the system is inherently extensible to real-world graphics design environments such as PowerPoint or Google Slides via APIs, and can incorporate additional computer graphics utilities—e.g., auto-layout, snapping, arrow libraries, style transfer—to expand aesthetic control while maintaining semantic correctness. As VLMs and LLMs continue to advance, the same architecture will naturally improve without retraining, offering increasing accuracy and efficiency. In current experiments, the primary bottleneck lies in the Critic VLM, which occasionally produces imprecise or inappropriate suggestions that can lead to worse candidates; this is mitigated by the Judge VLM, which filters regressions and ensures optimization stability. Beyond this application, VLM-based judging \cite{zheng_2023_judging} represents a promising approach for broader agentic systems and reinforcement learning in visual domains, but systematic evaluations of judge reliability—covering calibration, robustness, and domain generalization—are needed. Looking ahead, the qualitative-critique $\rightarrow$ multi-strategy synthesis $\rightarrow$ judging loop underlying our method could extend naturally to 3D tasks such as part assembly, scene layout \cite{liu_2025_ir3dbench}, and CAD editing \cite{jones_2021_automate}, leveraging recent advances in 3D spatial reasoning.

\newpage
\bibliographystyle{plain}
\sloppy
\bibliography{refs}

\begin{thebibliography}{19}
\providecommand{\natexlab}[1]{#1}
\providecommand{\url}[1]{\texttt{#1}}
\expandafter\ifx\csname urlstyle\endcsname\relax
  \providecommand{\doi}[1]{doi: #1}\else
  \providecommand{\doi}{doi: \begingroup \urlstyle{rm}\Url}\fi

\bibitem[Cai et~al.()Cai, Huang, Li, Ojha, Wang, and Lee]{cai_an}
Mu~Cai, Zeyi Huang, Yuheng Li, Utkarsh Ojha, Haohan Wang, and Yong Lee.
\newblock An investigation on llms' visual understanding ability using svg for image-text bridging.
\newblock URL \url{https://openaccess.thecvf.com/content/WACV2025/papers/Cai_An_Investigation_on_LLMs_Visual_Understanding_Ability_using_SVG_for_WACV_2025_paper.pdf}.

\bibitem[Carlier et~al.(2020)Carlier, Danelljan, Alahi, and Timofte]{carlier_2020_deepsvg}
Alexandre Carlier, Martin Danelljan, Alexandre Alahi, and Radu Timofte.
\newblock Deepsvg: A hierarchical generative network for vector graphics animation, 2020.
\newblock URL \url{https://arxiv.org/abs/2007.11301}.

\bibitem[Chen et~al.(2024)Chen, Xu, Kirmani, Ichter, Driess, Florence, Sadigh, Guibas, and Xia]{chen_2024_spatialvlm}
Boyuan Chen, Zhuo Xu, Sean Kirmani, Brian Ichter, Danny Driess, Pete Florence, Dorsa Sadigh, Leonidas Guibas, and Fei Xia.
\newblock Spatialvlm: Endowing vision-language models with spatial reasoning capabilities, 2024.
\newblock URL \url{https://arxiv.org/abs/2401.12168}.

\bibitem[Gu et~al.(2025)Gu, Huang, Je, Yang, and Guibas]{gu_2025_blendergym}
Yunqi Gu, Ian Huang, Jihyeon Je, Guandao Yang, and Leonidas Guibas.
\newblock Blendergym: Benchmarking foundational model systems for graphics editing, 2025.
\newblock URL \url{https://arxiv.org/abs/2504.01786}.

\bibitem[Hu et~al.(2024)Hu, Iscen, Jain, Kipf, Yue, Ross, Schmid, and Fathi]{hu_2024_scenecraft}
Ziniu Hu, Ahmet Iscen, Aashi Jain, Thomas Kipf, Yisong Yue, David~A Ross, Cordelia Schmid, and Alireza Fathi.
\newblock Scenecraft: An llm agent for synthesizing 3d scene as blender code, 2024.
\newblock URL \url{https://arxiv.org/abs/2403.01248}.

\bibitem[Huang et~al.(2024)Huang, Yang, and Guibas]{huang_2024_blenderalchemy}
Ian Huang, Guandao Yang, and Leonidas Guibas.
\newblock Blenderalchemy: Editing 3d graphics with vision-language models, 2024.
\newblock URL \url{https://arxiv.org/abs/2404.17672}.

\bibitem[Jones et~al.(2021)Jones, Hildreth, Chen, Baran, Kim, and Schulz]{jones_2021_automate}
Benjamin Jones, Dalton Hildreth, Duowen Chen, Ilya Baran, Vladimir~G. Kim, and Adriana Schulz.
\newblock Automate.
\newblock \emph{ACM Transactions on Graphics}, 40:\penalty0 1--18, 12 2021.
\newblock \doi{10.1145/3478513.3480562}.

\bibitem[Kapur et~al.(2024)Kapur, Jenner, and Russell]{kapur_2024_diffusion}
Shreyas Kapur, Erik Jenner, and Stuart Russell.
\newblock Diffusion on syntax trees for program synthesis, 2024.
\newblock URL \url{https://arxiv.org/abs/2405.20519}.

\bibitem[Kawar et~al.()Kawar, Zada, Lang, Tov, Chang, Dekel, Mosseri, and Irani]{kawar_imagic}
Bahjat Kawar, Shiran Zada, Oran Lang, Omer Tov, Huiwen Chang, Tali Dekel, Inbar Mosseri, and Michal Irani.
\newblock Imagic: Text-based real image editing with diffusion models.
\newblock URL \url{https://openaccess.thecvf.com/content/CVPR2023/papers/Kawar_Imagic_Text-Based_Real_Image_Editing_With_Diffusion_Models_CVPR_2023_paper.pdf}.

\bibitem[Koley et~al.(2023)Koley, Kumar, Sain, Chowdhury, Xiang, and Song]{koley_2023_picture}
Subhadeep Koley, Bhunia~Ayan Kumar, Aneeshan Sain, Pinaki~Nath Chowdhury, Tao Xiang, and Yi-Zhe Song.
\newblock Picture that sketch: Photorealistic image generation from abstract sketches, 2023.
\newblock URL \url{https://arxiv.org/abs/2303.11162}.

\bibitem[Liu et~al.(2025)Liu, Li, Li, Wu, Li, Yang, Zhang, Lin, Han, and Feng]{liu_2025_ir3dbench}
Parker Liu, Chenxin Li, Zhengxin Li, Yipeng Wu, Wuyang Li, Zhiqin Yang, Zhenyuan Zhang, Yunlong Lin, Sirui Han, and Brandon~Y Feng.
\newblock Ir3d-bench: Evaluating vision-language model scene understanding as agentic inverse rendering, 2025.
\newblock URL \url{https://arxiv.org/abs/2506.23329}.

\bibitem[Mou et~al.(2023)Mou, Wang, Xie, Wu, Zhang, Qi, Shan, and Qie]{mou_2023_t2iadapter}
Chong Mou, Xintao Wang, Liangbin Xie, Yanze Wu, Jian Zhang, Zhongang Qi, Ying Shan, and Xiaohu Qie.
\newblock T2i-adapter: Learning adapters to dig out more controllable ability for text-to-image diffusion models.
\newblock \emph{arXiv:2302.08453 [cs]}, 03 2023.
\newblock URL \url{https://arxiv.org/abs/2302.08453}.

\bibitem[Wu et~al.()Wu, Su, and Liao]{wu_chat2svg}
Ronghuan Wu, Wanchao Su, and Jing Liao.
\newblock Chat2svg: Vector graphics generation with large language models and image diffusion models.
\newblock URL \url{https://openaccess.thecvf.com/content/CVPR2025/papers/Wu_Chat2SVG_Vector_Graphics_Generation_with_Large_Language_Models_and_Image_CVPR_2025_paper.pdf}.

\bibitem[Wu et~al.(2024)Wu, Mao, Zhang, Xia, Dong, Cui, and Wei]{wu_2024_visualizationofthought}
Wenshan Wu, Shaoguang Mao, Yadong Zhang, Yan Xia, Li~Dong, Lei Cui, and Furu Wei.
\newblock Visualization-of-thought elicits spatial reasoning in large language models, 04 2024.
\newblock URL \url{https://arxiv.org/abs/2404.03622}.

\bibitem[Yamada et~al.(2023)Yamada, Bao, Lampinen, Kasai, and Yildirim]{yamada_2023_evaluating}
Yutaro Yamada, Yihan Bao, Andrew~K Lampinen, Jungo Kasai, and Ilker Yildirim.
\newblock Evaluating spatial understanding of large language models, 2023.
\newblock URL \url{https://arxiv.org/abs/2310.14540}.

\bibitem[Yamada et~al.(2024)Yamada, Chandu, Lin, Hessel, Yildirim, and Choi]{yamada_2024_l3go}
Yutaro Yamada, Khyathi Chandu, Yuchen Lin, Jack Hessel, Ilker Yildirim, and Yejin Choi.
\newblock L3go: Language agents with chain-of-3d-thoughts for generating unconventional objects, 2024.
\newblock URL \url{https://arxiv.org/abs/2402.09052}.

\bibitem[Zhang and Agrawala(2023)]{zhang_2023_adding}
Lvmin Zhang and Maneesh Agrawala.
\newblock Adding conditional control to text-to-image diffusion models, 02 2023.
\newblock URL \url{https://arxiv.org/abs/2302.05543}.

\bibitem[Zhang et~al.(2024)Zhang, Xie, Du, and Xie]{zhang_2024_sketchguided}
Tianyu Zhang, Xiaoxuan Xie, Xusheng Du, and Haoran Xie.
\newblock Sketch-guided scene image generation, 2024.
\newblock URL \url{https://arxiv.org/abs/2407.06469}.

\bibitem[Zheng et~al.(2023)Zheng, Chiang, Sheng, Zhuang, Wu, Zhuang, Lin, Li, Li, Xing, Zhang, Gonzalez, and Stoica]{zheng_2023_judging}
Lianmin Zheng, Wei-Lin Chiang, Ying Sheng, Siyuan Zhuang, Zhanghao Wu, Yonghao Zhuang, Zi~Lin, Zhuohan Li, Dacheng Li, Eric~P. Xing, Hao Zhang, Joseph~E. Gonzalez, and Ion Stoica.
\newblock Judging llm-as-a-judge with mt-bench and chatbot arena, 07 2023.
\newblock URL \url{https://arxiv.org/abs/2306.05685}.

\end{thebibliography}

\newpage
\appendix

\section{Appendix}

\subsection{Comparison with Frontier Models}
\begin{table}[ht]
\centering
\renewcommand{\arraystretch}{1.2} 
\setlength{\tabcolsep}{6pt}       

\begin{tabular}{|m{0.22\textwidth}|m{0.22\textwidth}|m{0.22\textwidth}|m{0.22\textwidth}|}
\hline
\textbf{Sketch} & \textbf{See it. Say it. Sorted.} & \textbf{GPT-5} & \textbf{Gemini-2.5-Pro} \\ \hline

\includegraphics[width=\linewidth]{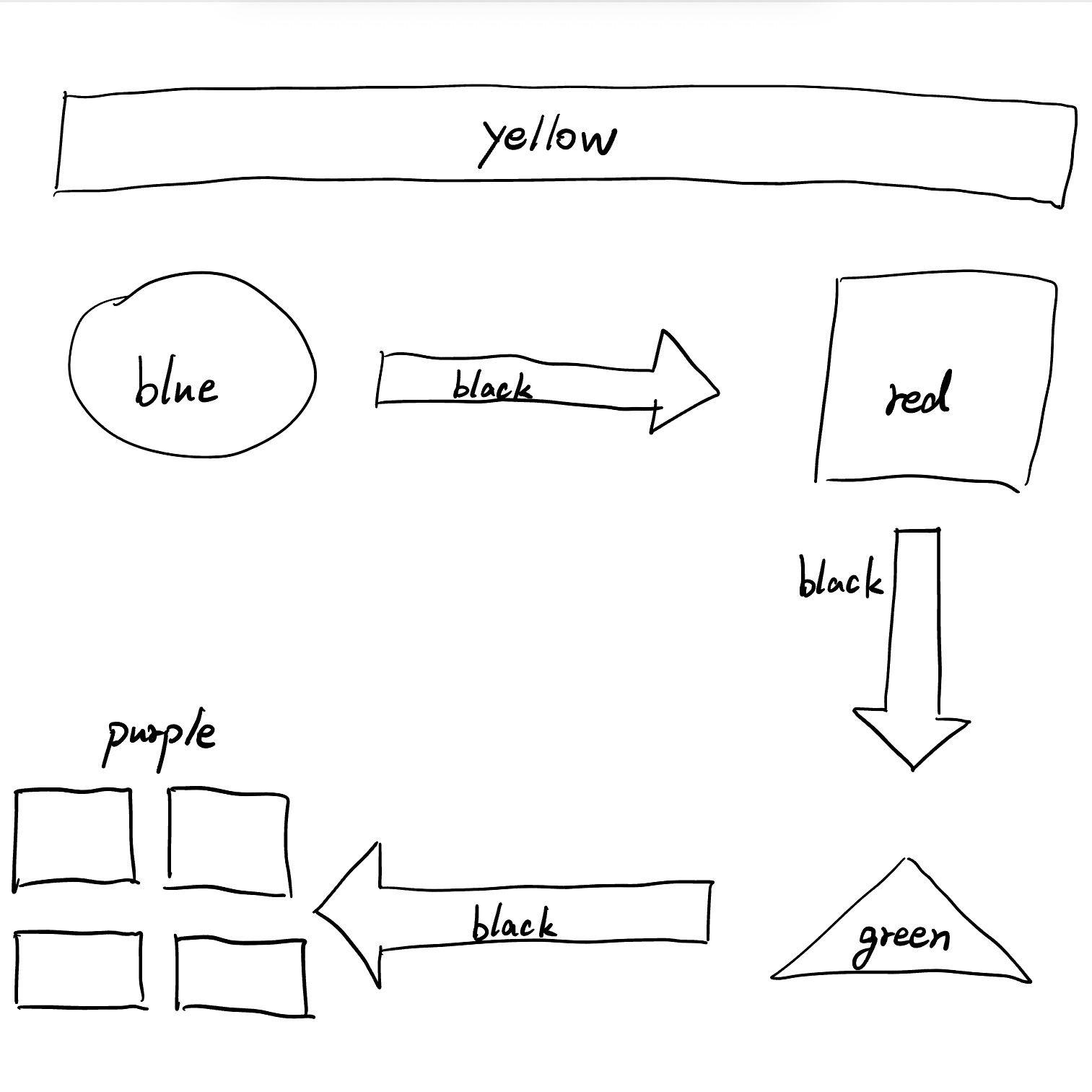} &
\includegraphics[width=\linewidth]{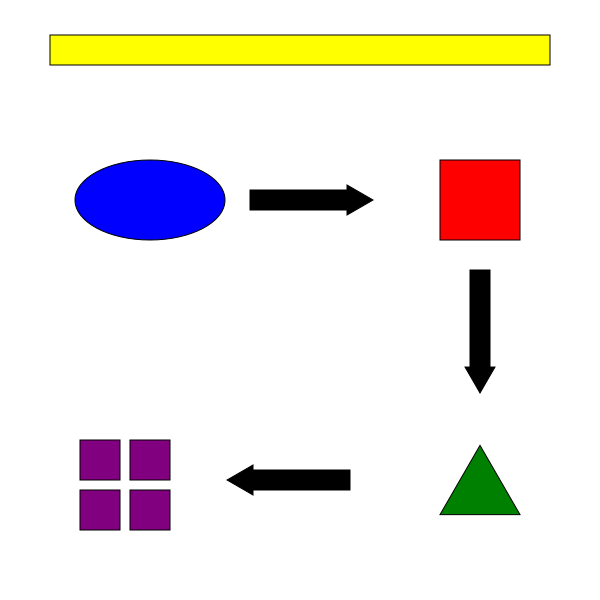} &
\includegraphics[width=\linewidth]{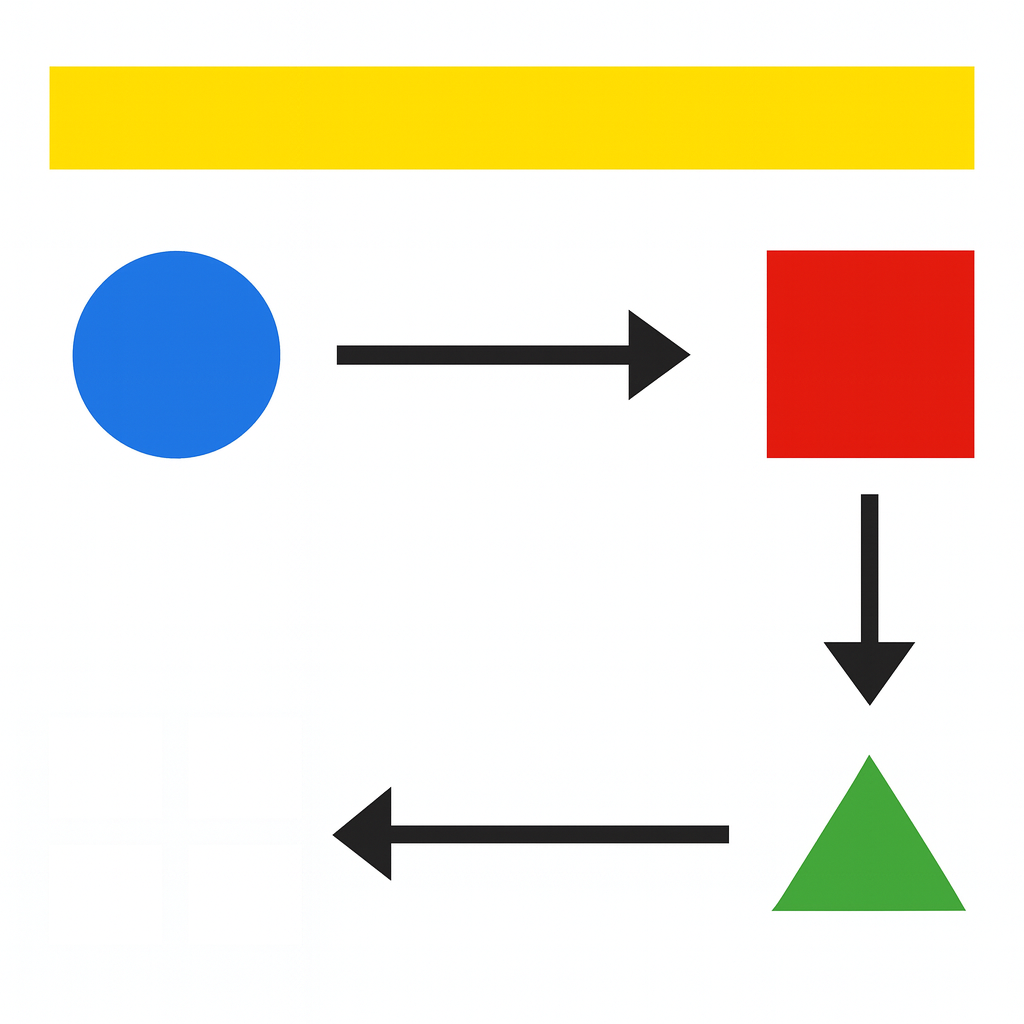} &
\includegraphics[width=\linewidth]{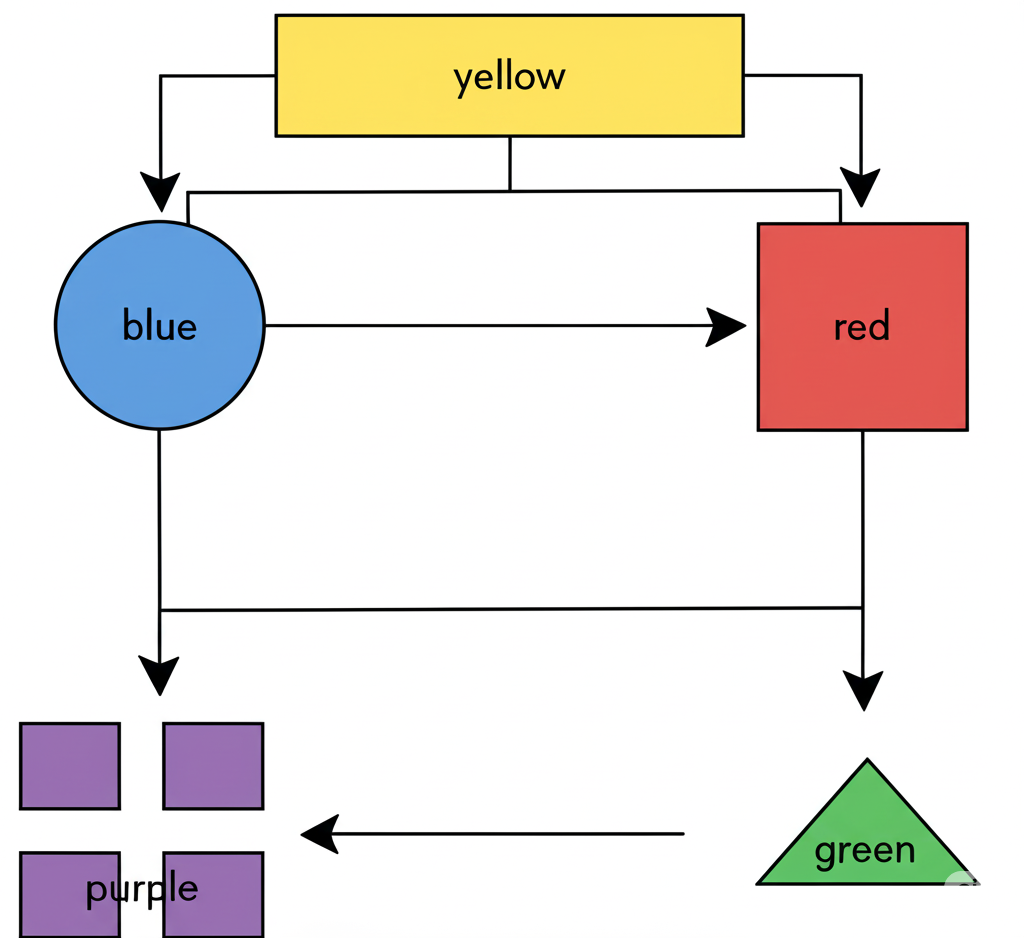} \\ \hline

\includegraphics[width=\linewidth]{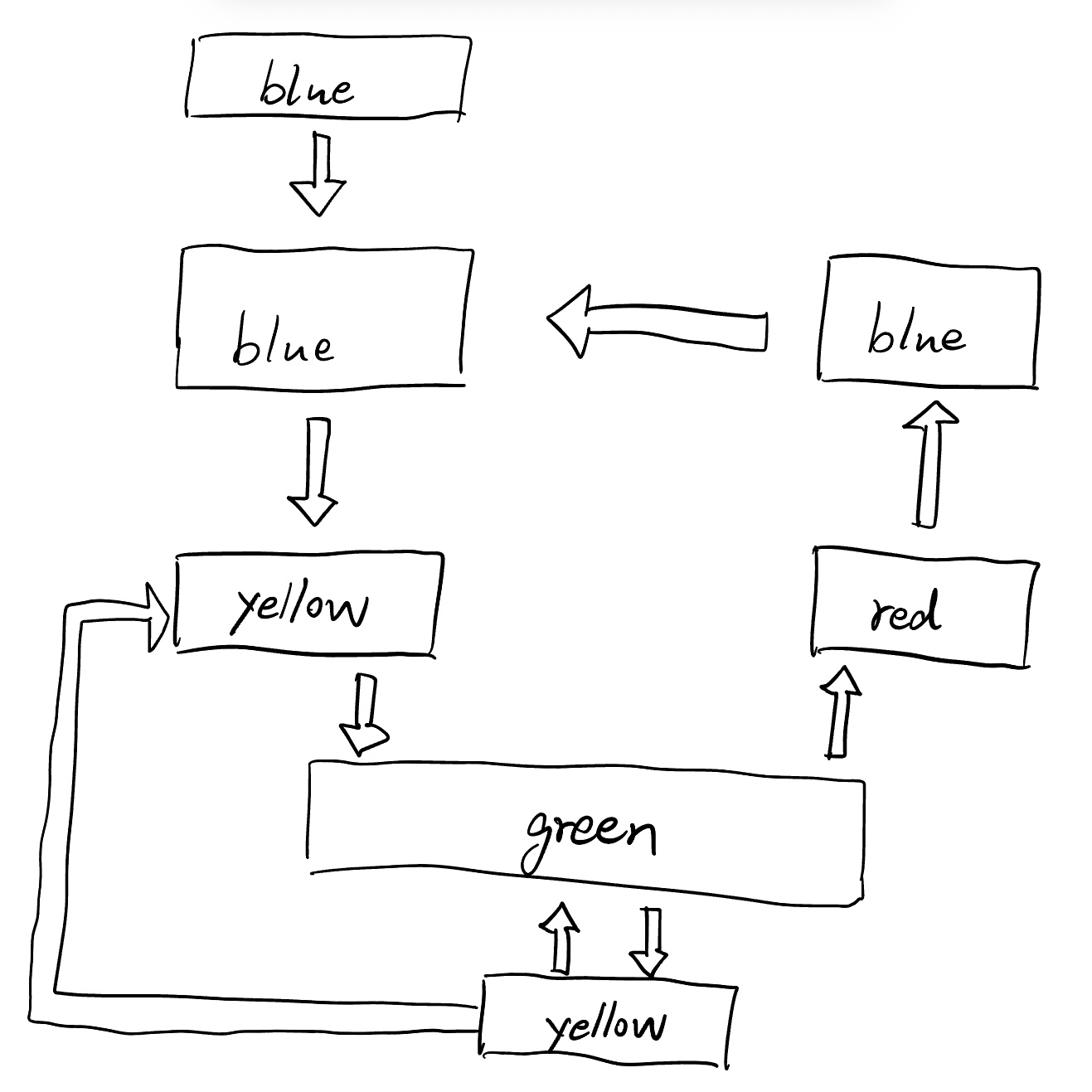} &
\includegraphics[width=\linewidth]{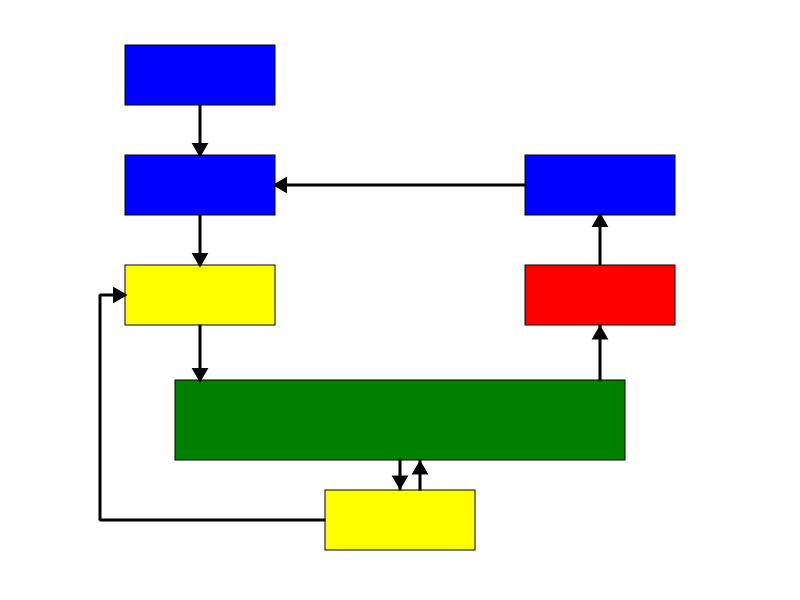} &
\includegraphics[width=\linewidth]{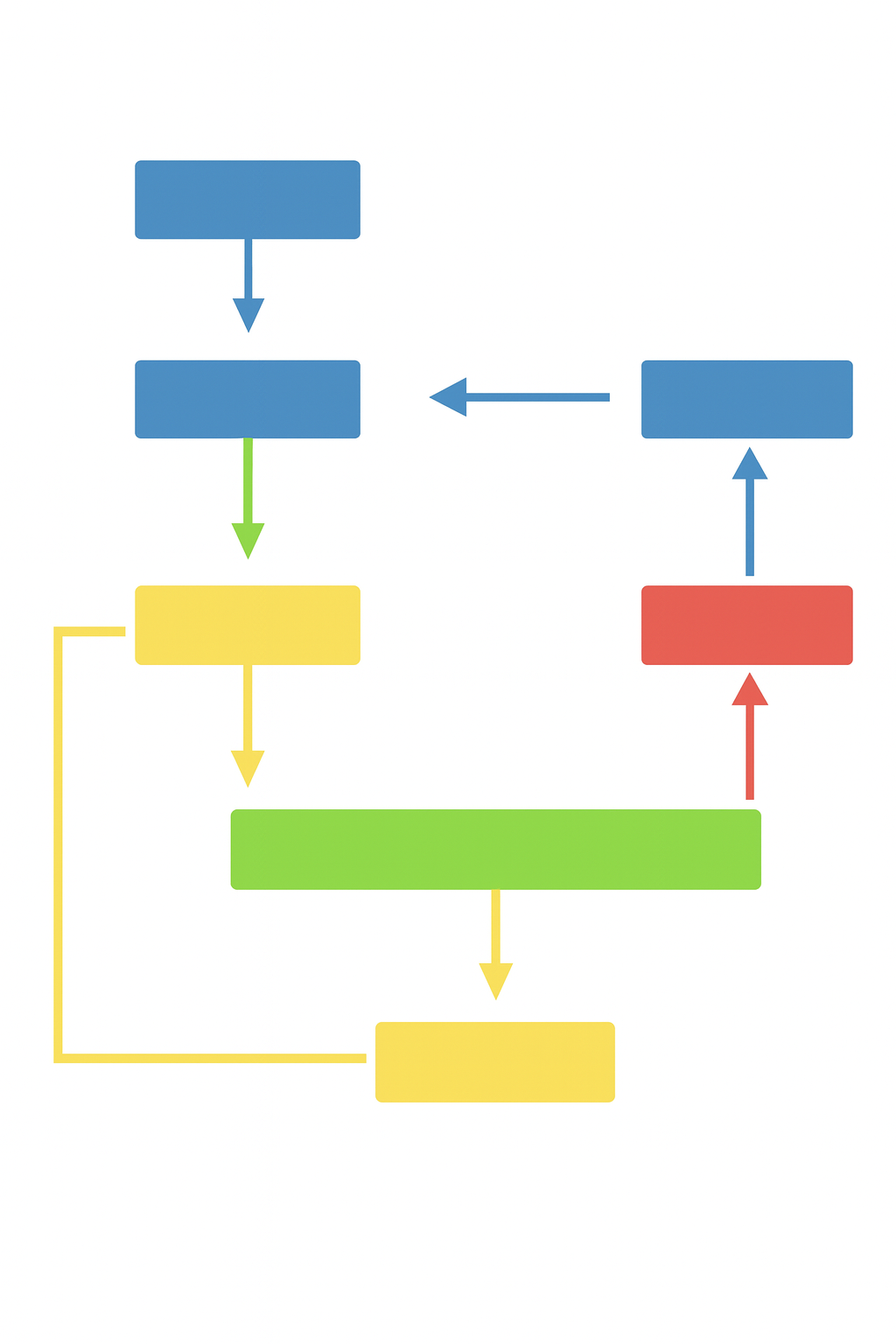} &
\includegraphics[width=\linewidth]{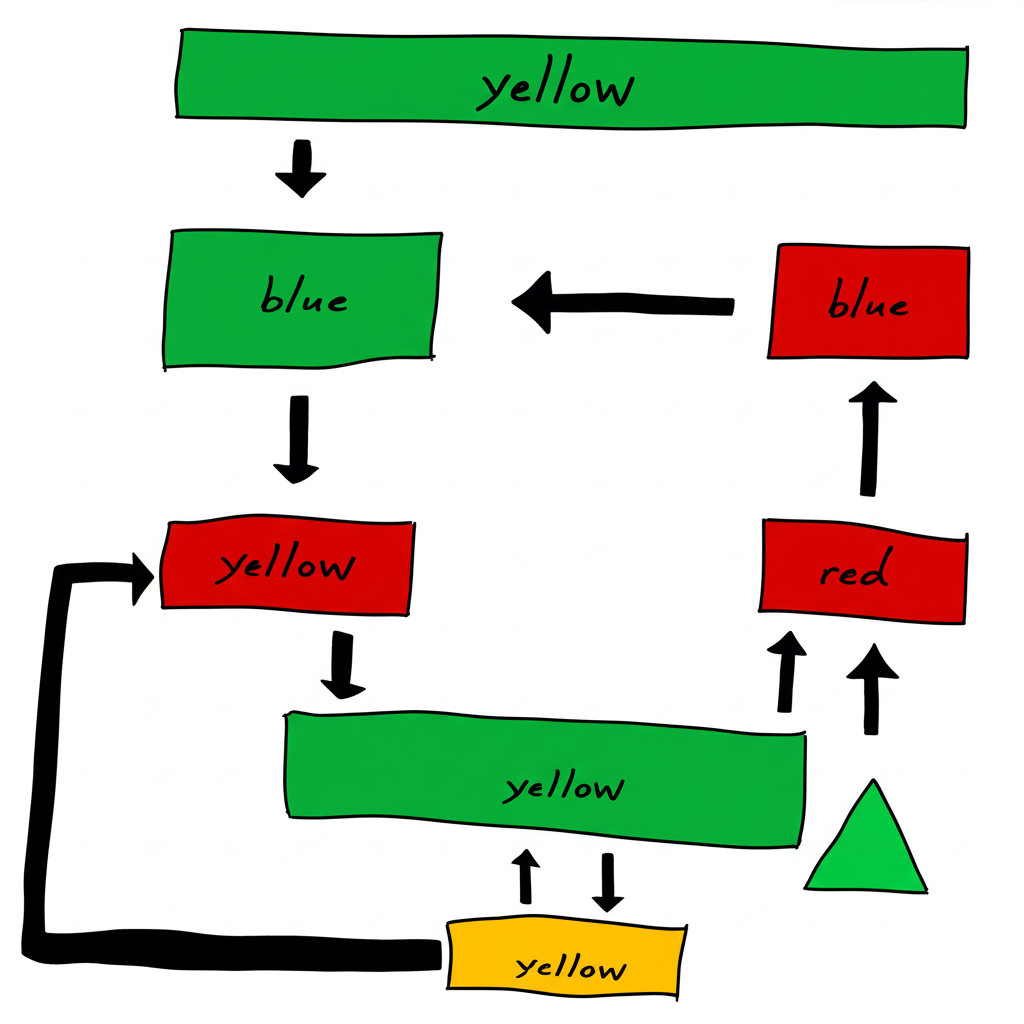} \\ \hline

\includegraphics[width=\linewidth]{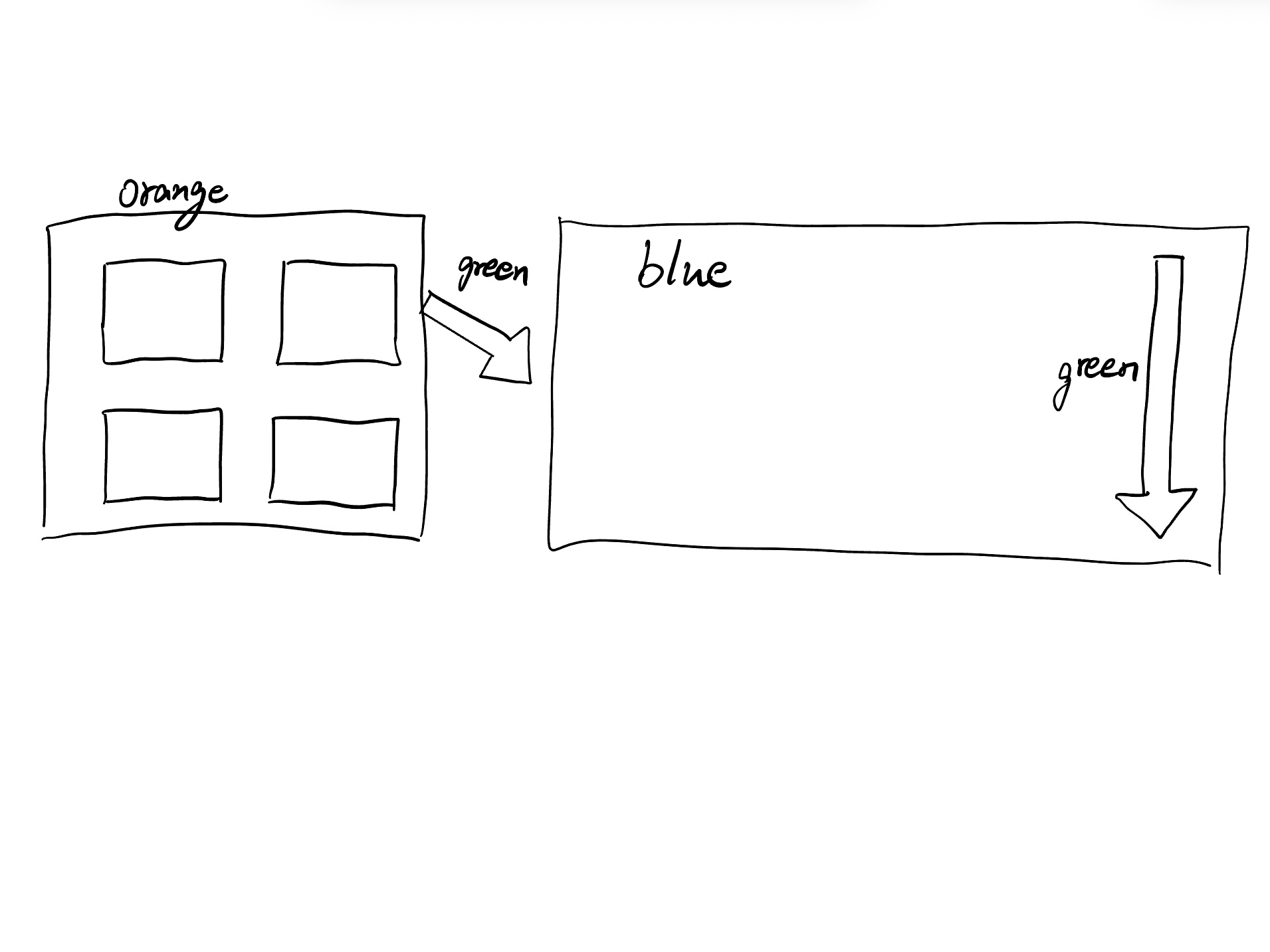} &
\includegraphics[width=\linewidth]{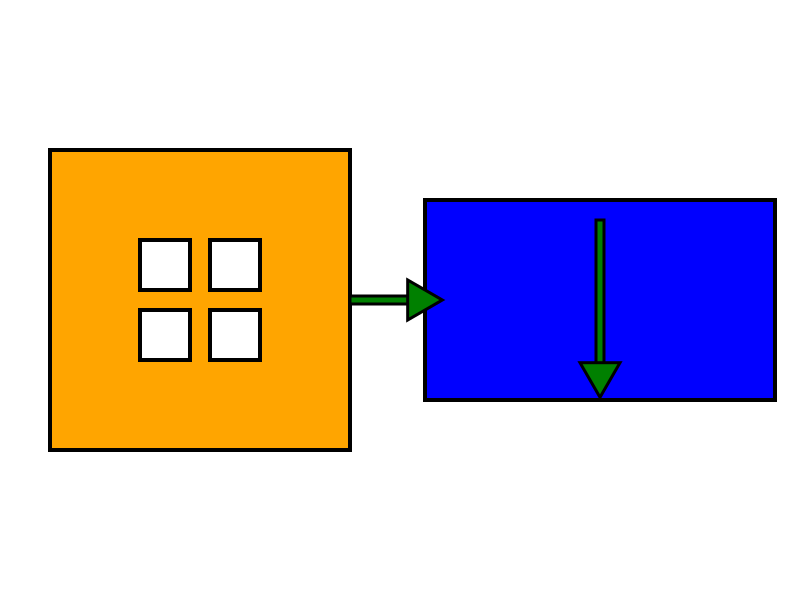} &
\includegraphics[width=\linewidth]{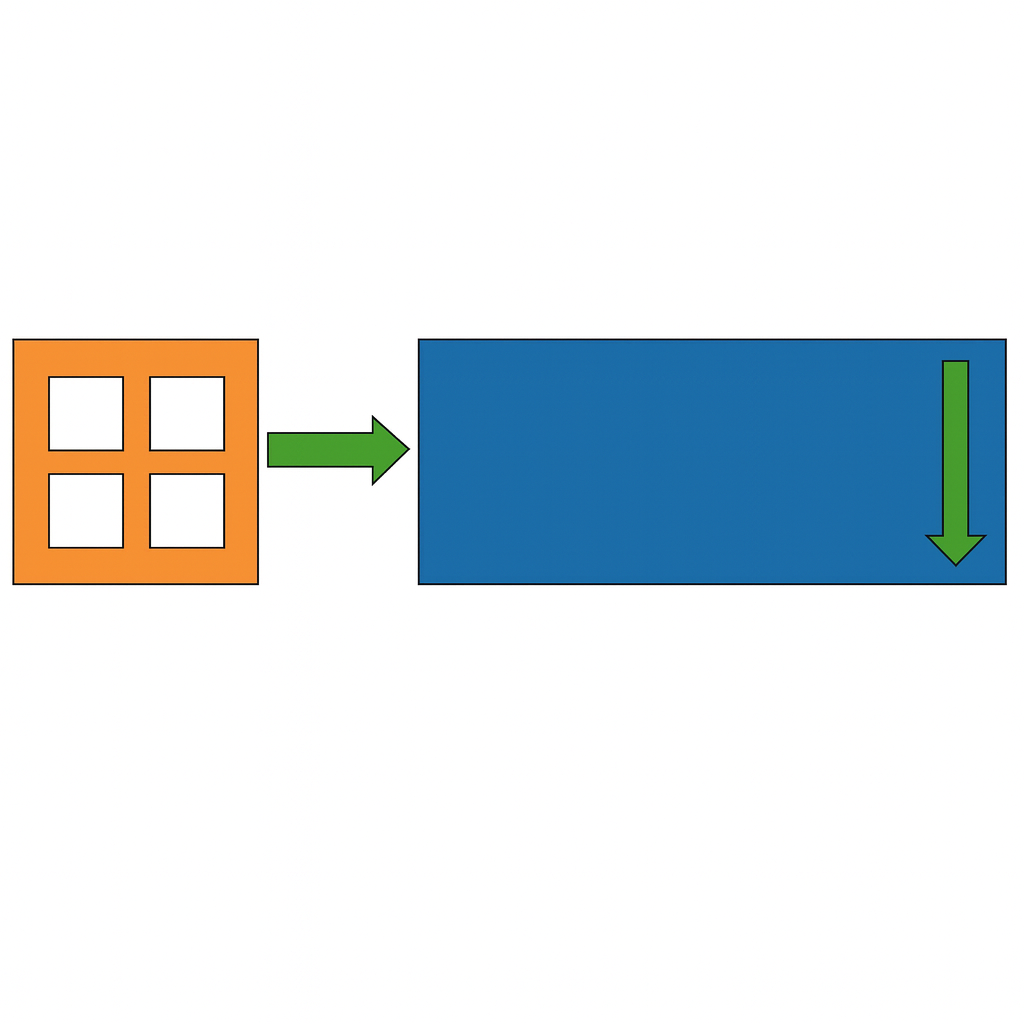} &
\includegraphics[width=\linewidth]{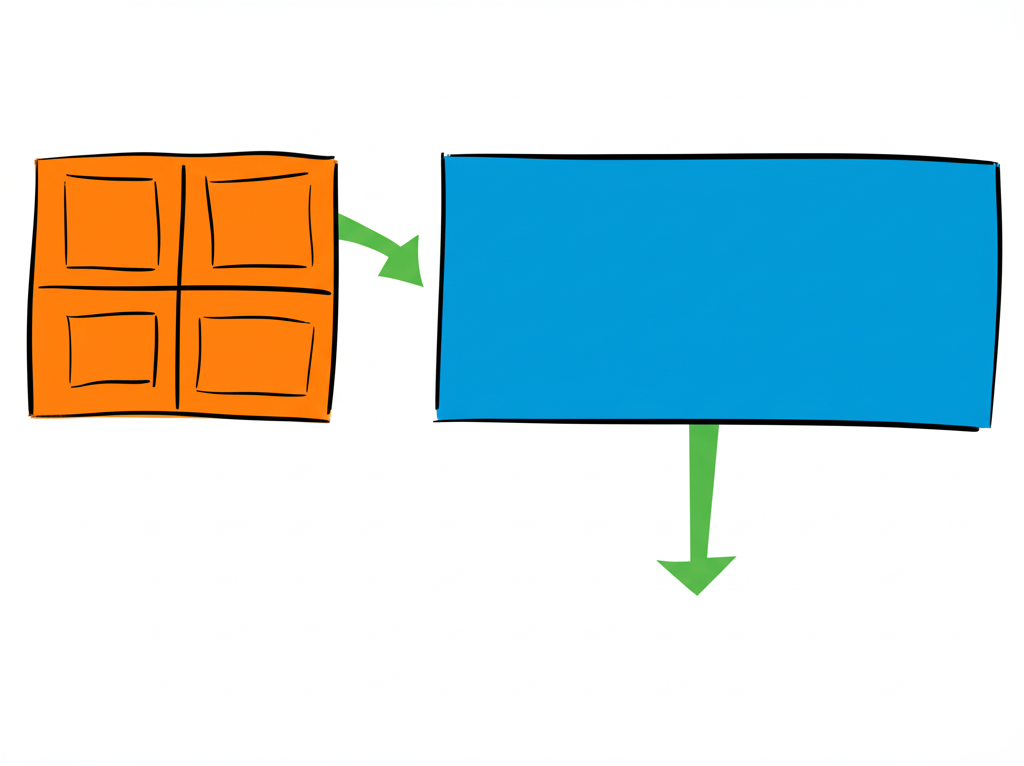} \\ \hline

\includegraphics[width=\linewidth]{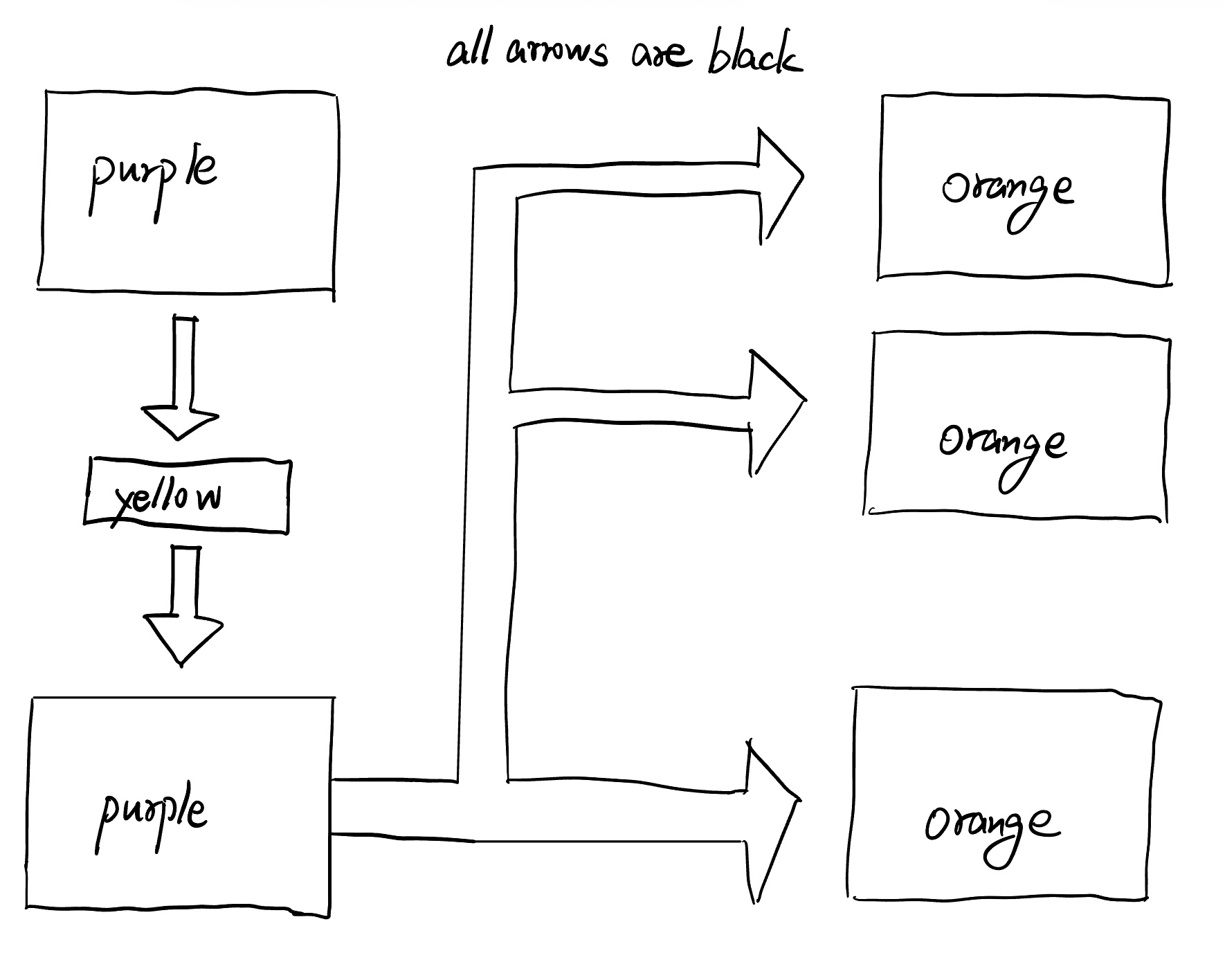} &
\includegraphics[width=\linewidth]{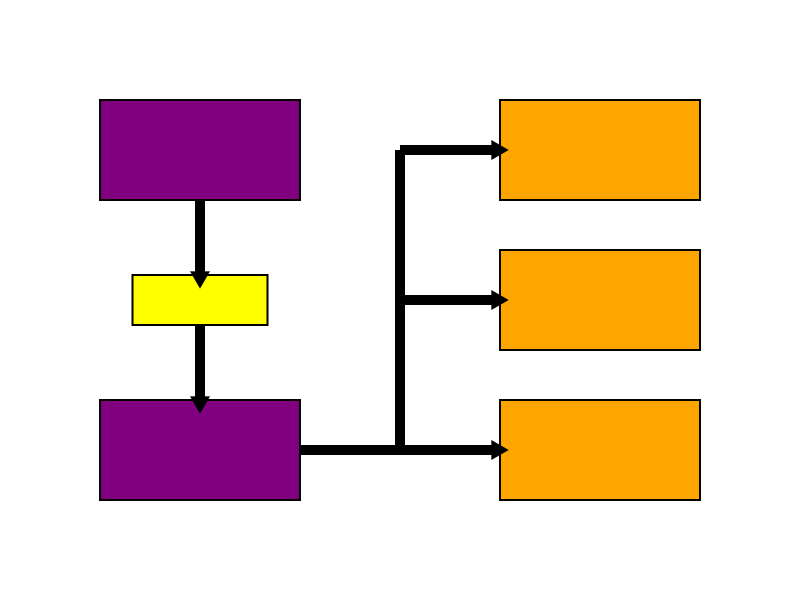} &
\includegraphics[width=\linewidth]{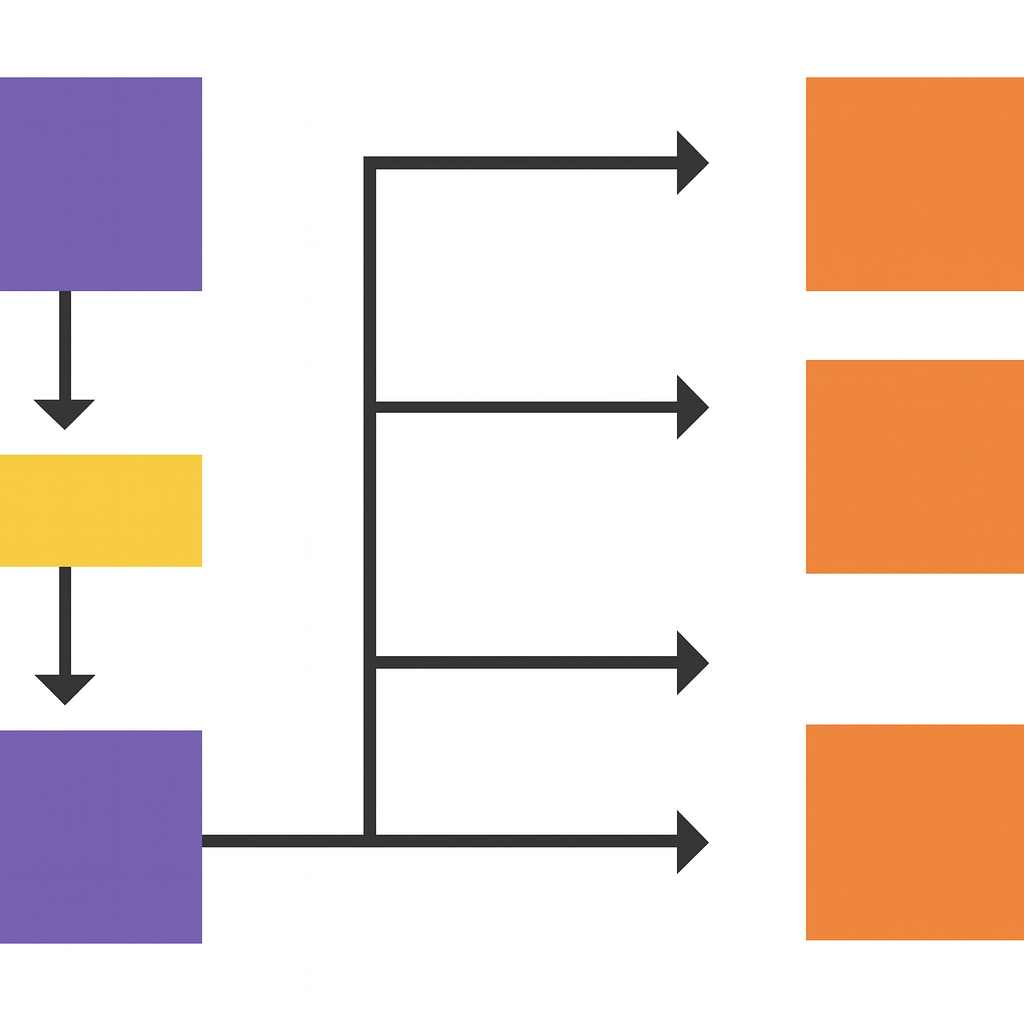} &
\includegraphics[width=\linewidth]{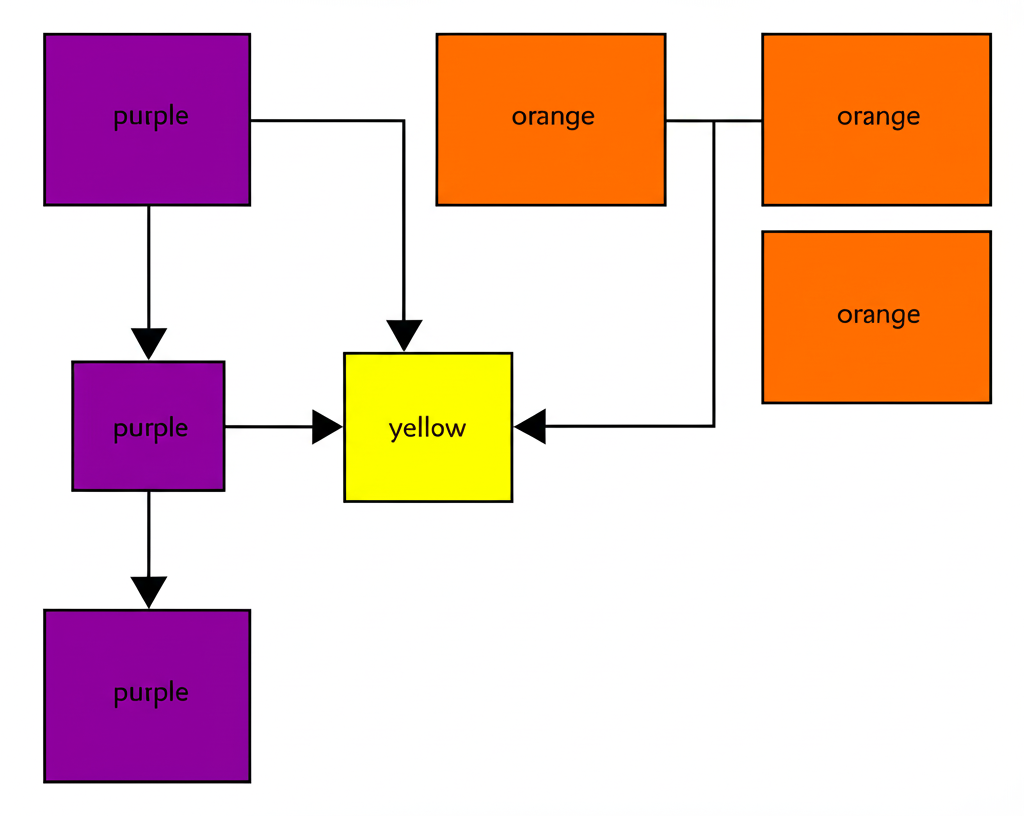} \\ \hline

\includegraphics[width=\linewidth]{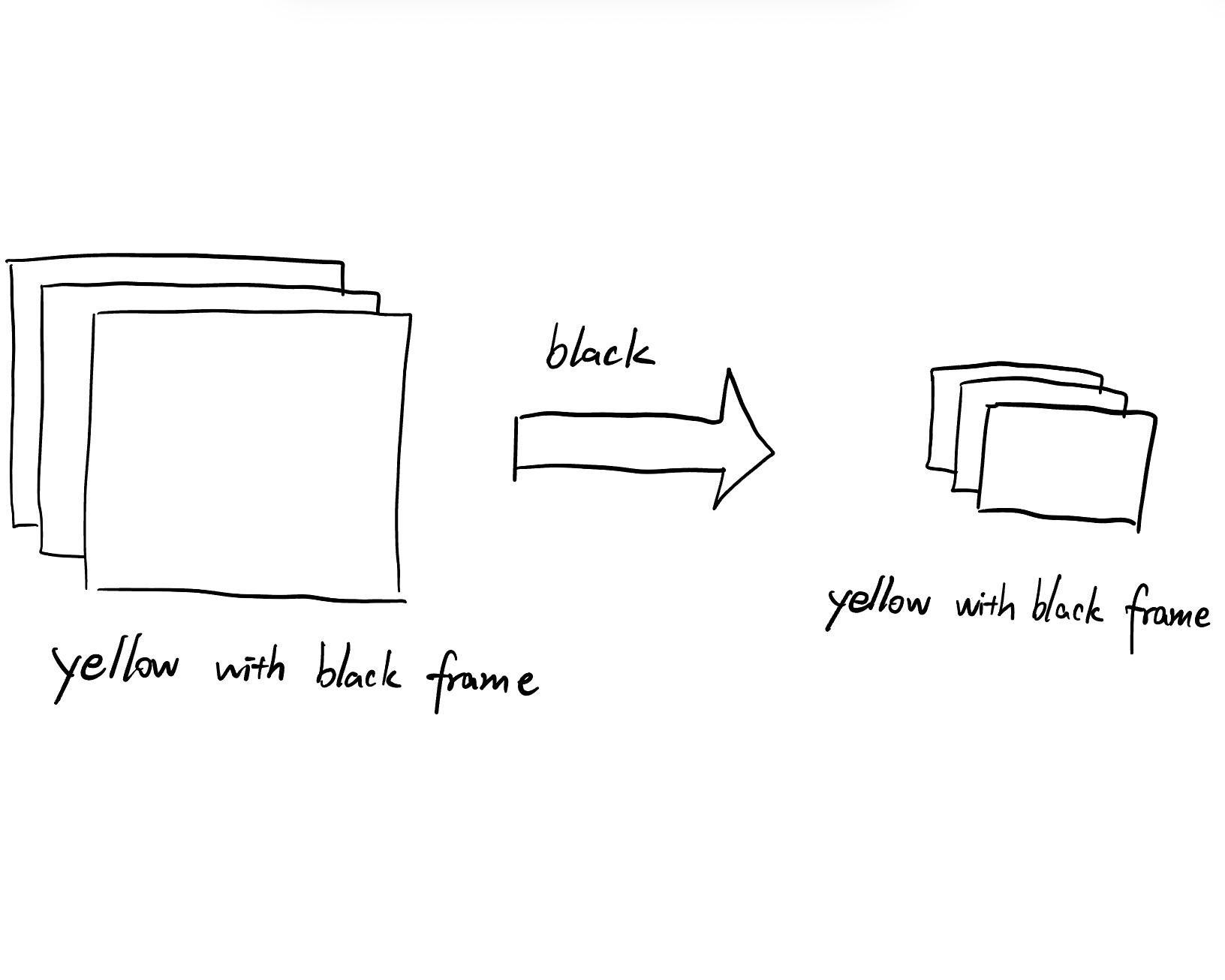} &
\includegraphics[width=\linewidth]{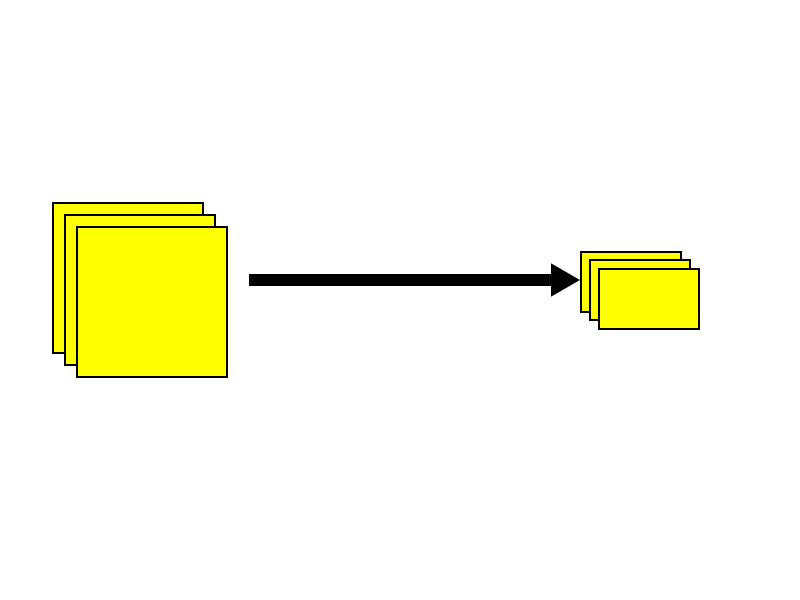} &
\includegraphics[width=\linewidth]{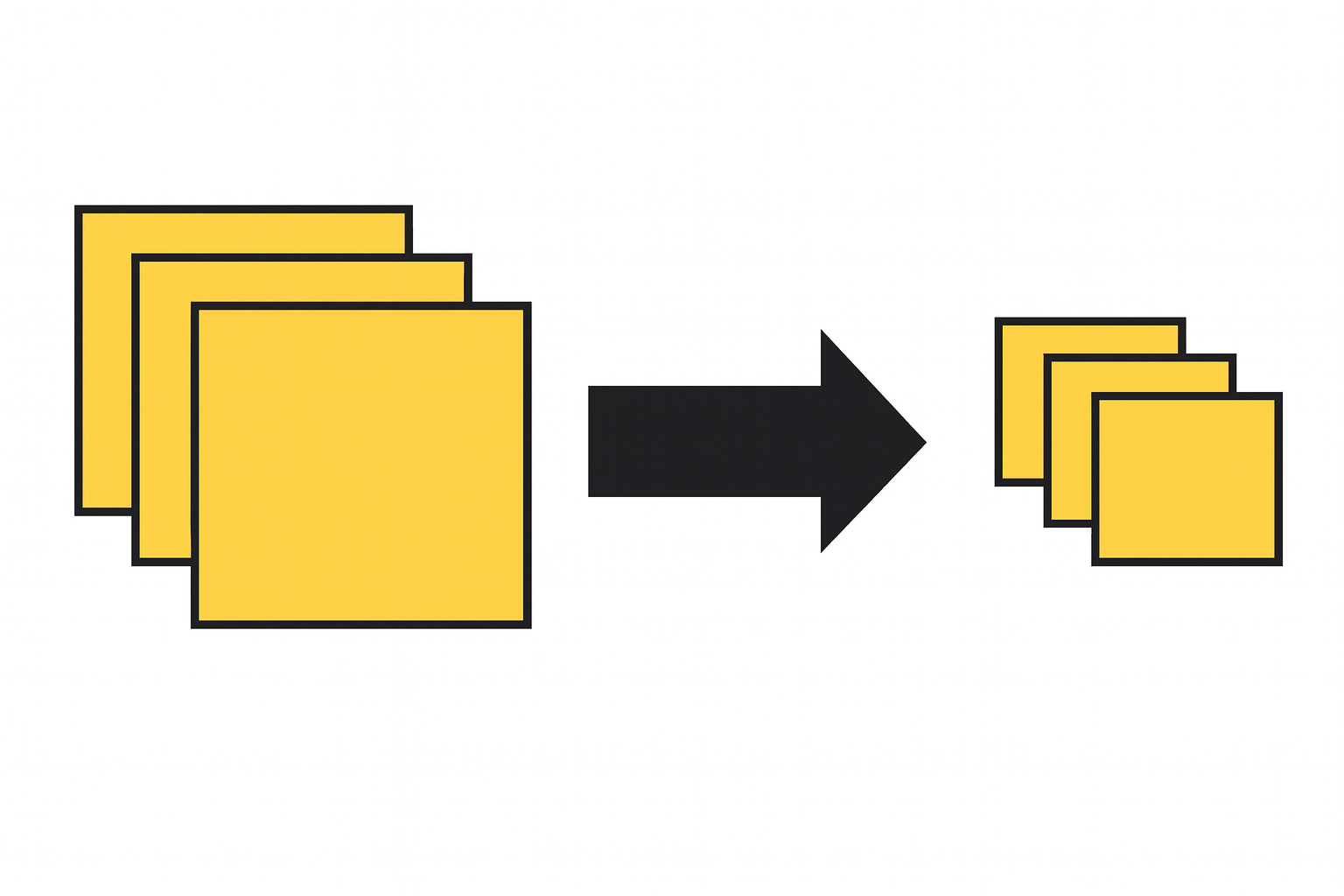} &
\includegraphics[width=\linewidth]{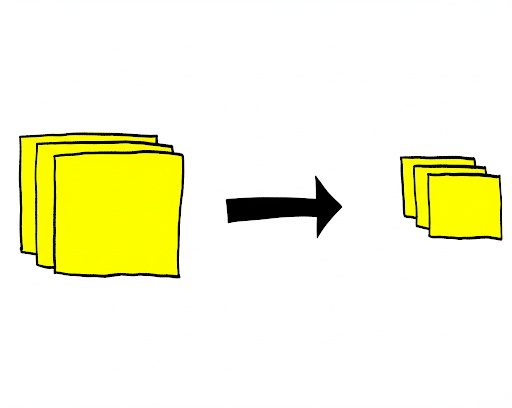} \\ \hline

\end{tabular}
\caption{Task 1-5}
\label{tab:comparison_1}
\end{table}

\newpage

\begin{table}[ht]
\centering
\renewcommand{\arraystretch}{1.2} 
\setlength{\tabcolsep}{6pt}       

\begin{tabular}{|m{0.22\textwidth}|m{0.22\textwidth}|m{0.22\textwidth}|m{0.22\textwidth}|}
\hline
\textbf{Sketch} & \textbf{See it. Say it. Sorted.} & \textbf{GPT-5} & \textbf{Gemini-2.5-Pro} \\ \hline

\includegraphics[width=\linewidth]{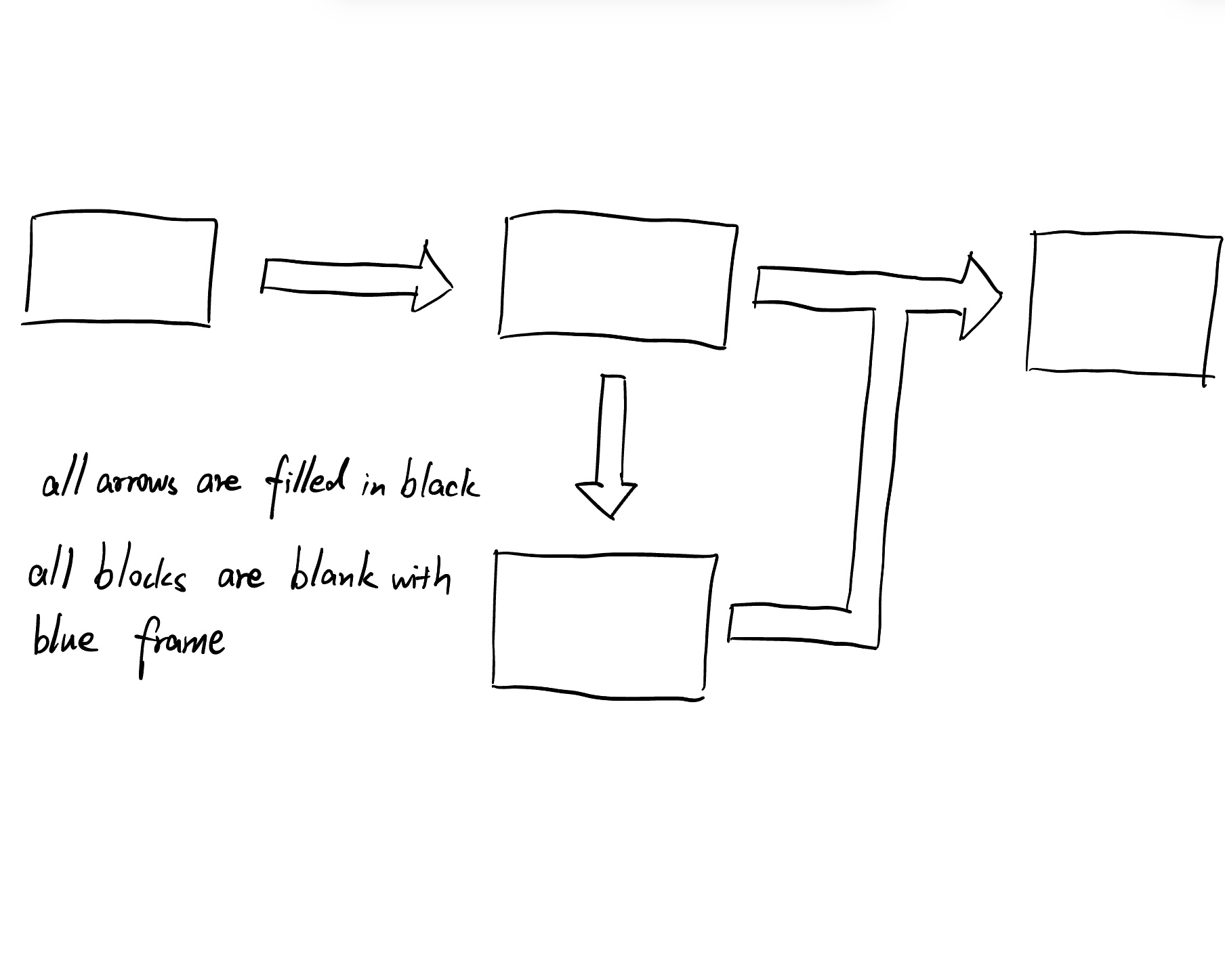} &
\includegraphics[width=\linewidth]{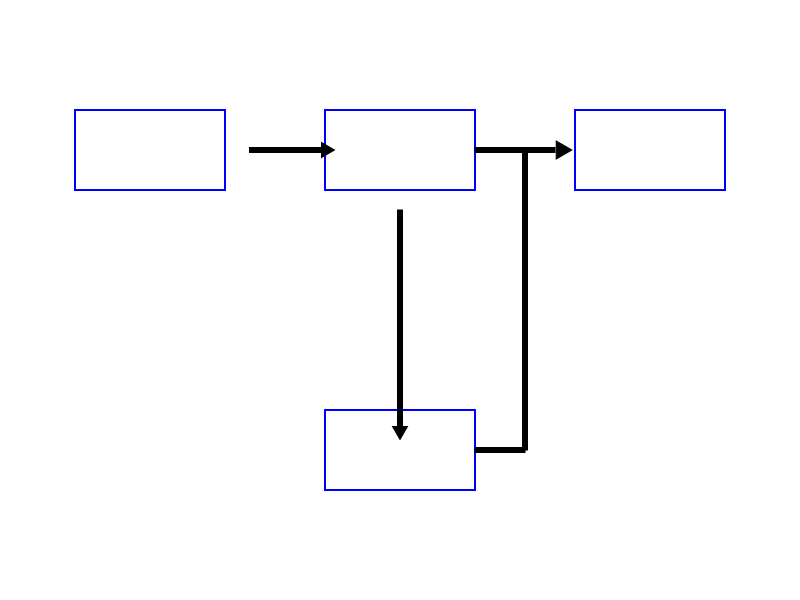} &
\includegraphics[width=\linewidth]{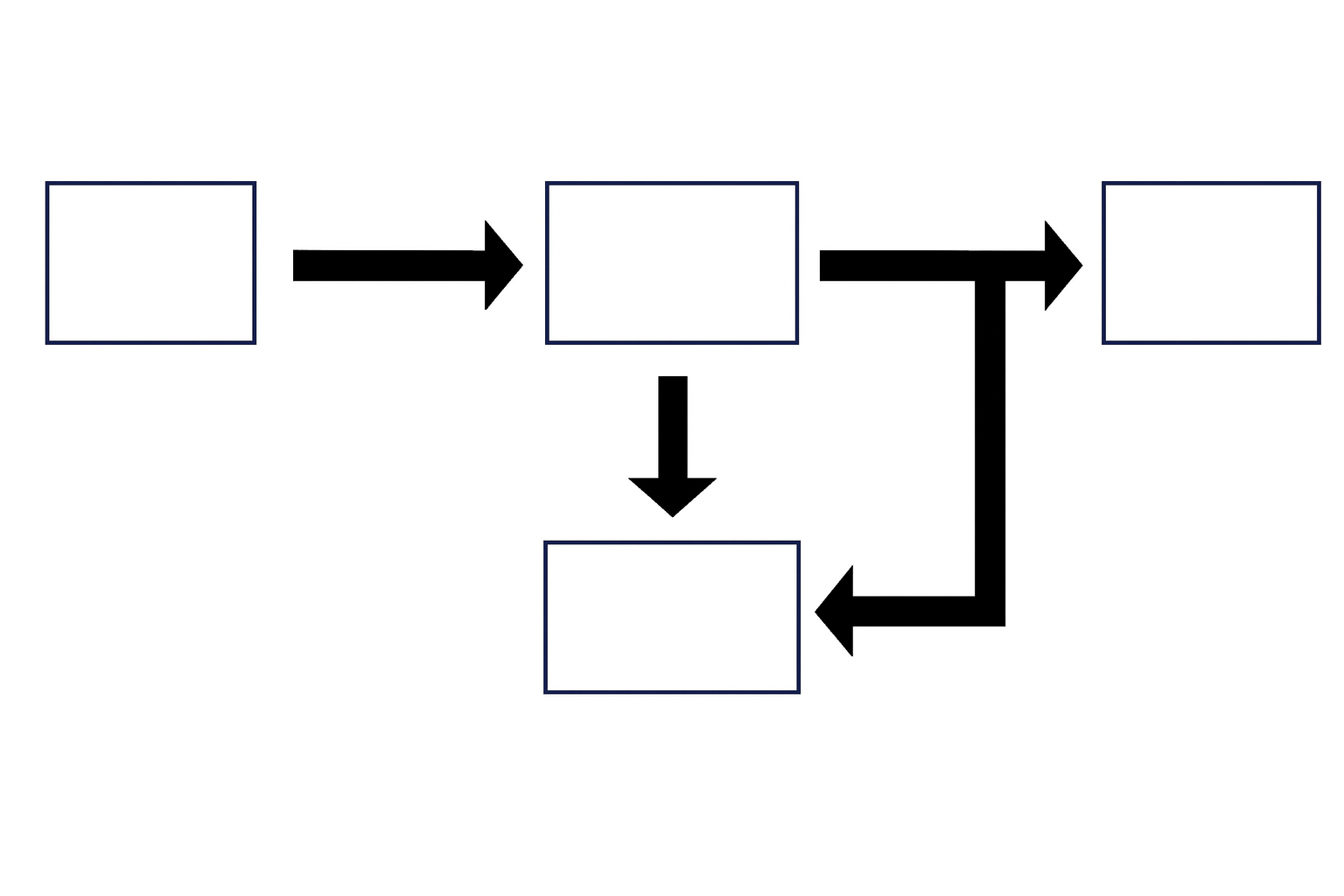} &
\includegraphics[width=\linewidth]{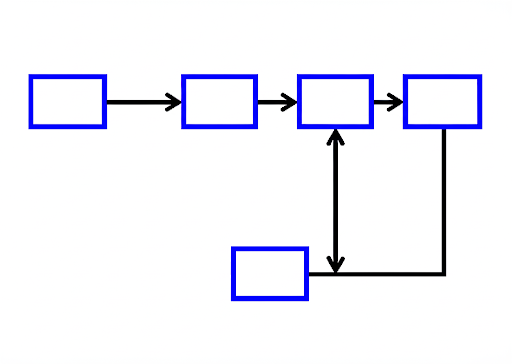} \\ \hline

\includegraphics[width=\linewidth]{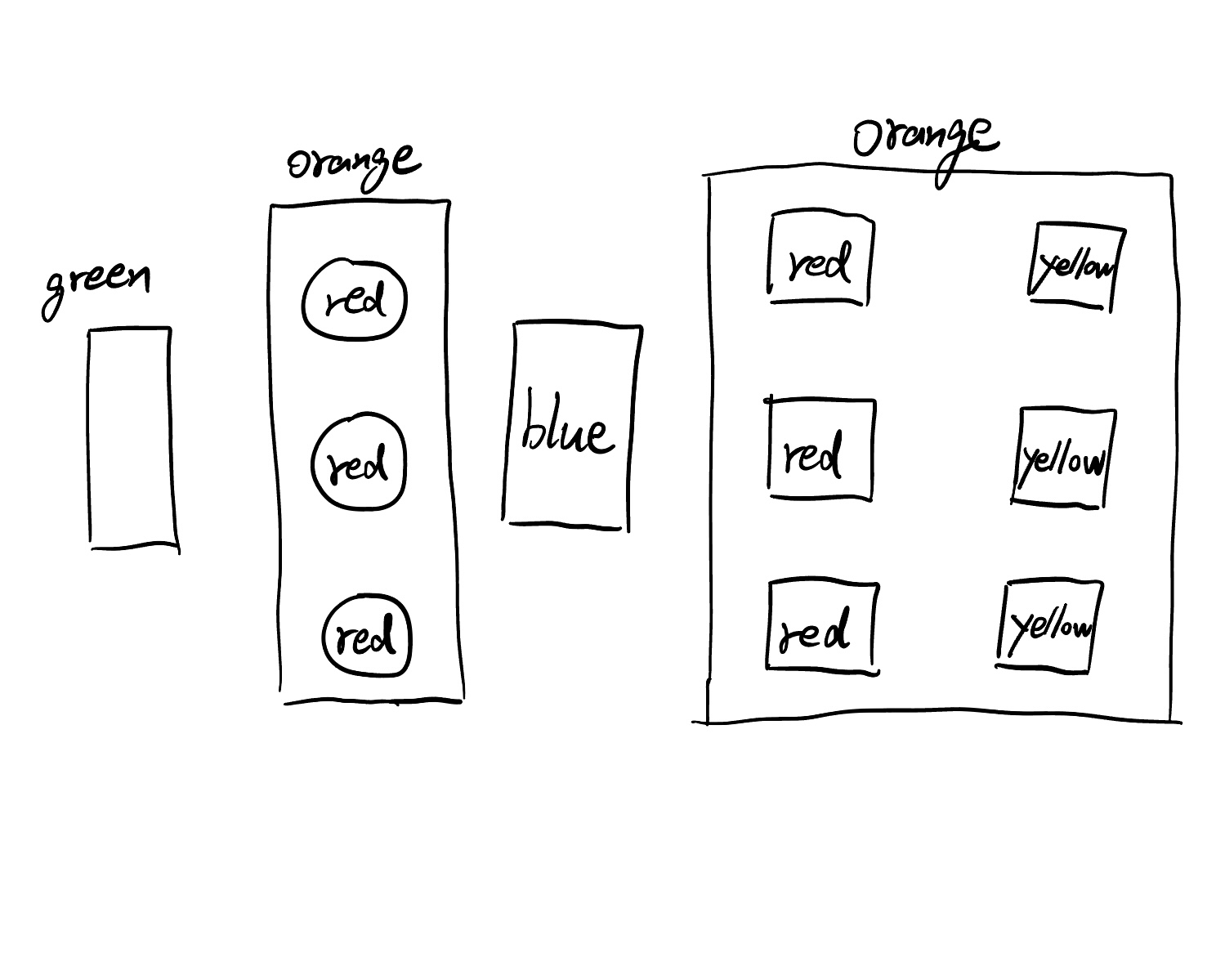} &
\includegraphics[width=\linewidth]{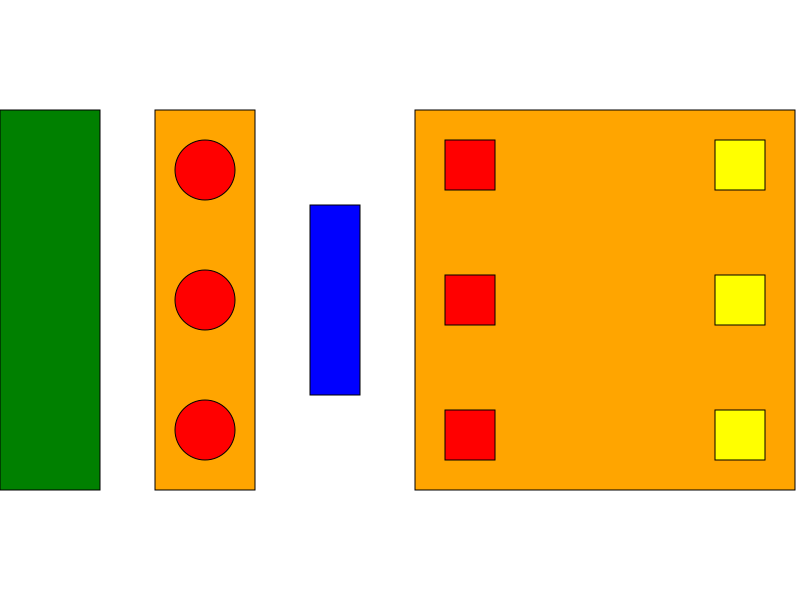} &
\includegraphics[width=\linewidth]{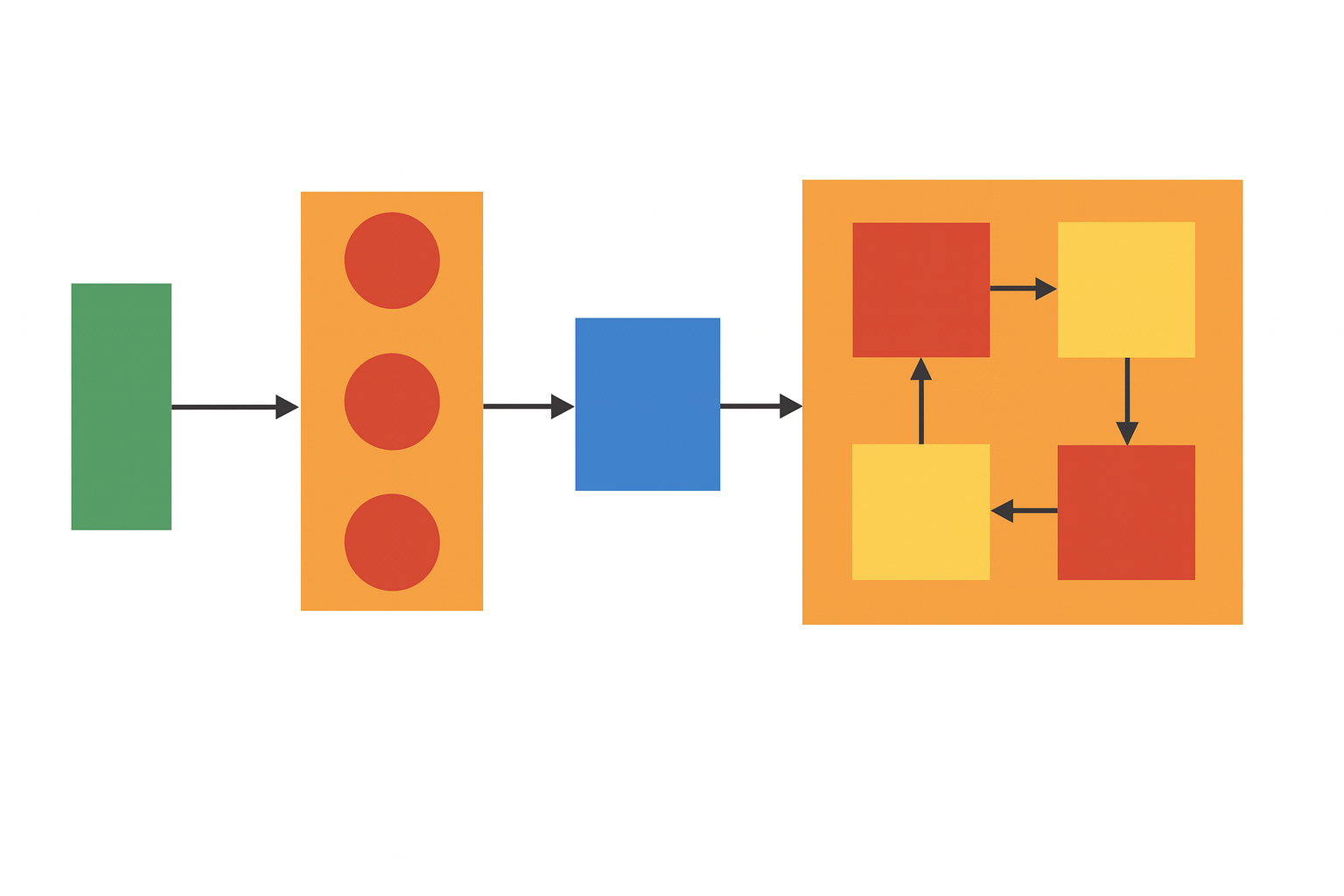} &
\includegraphics[width=\linewidth]{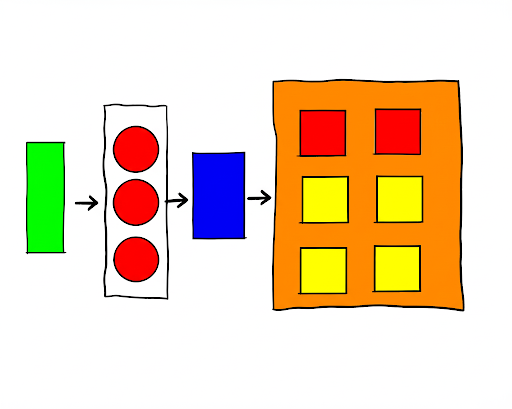} \\ \hline

\includegraphics[width=\linewidth]{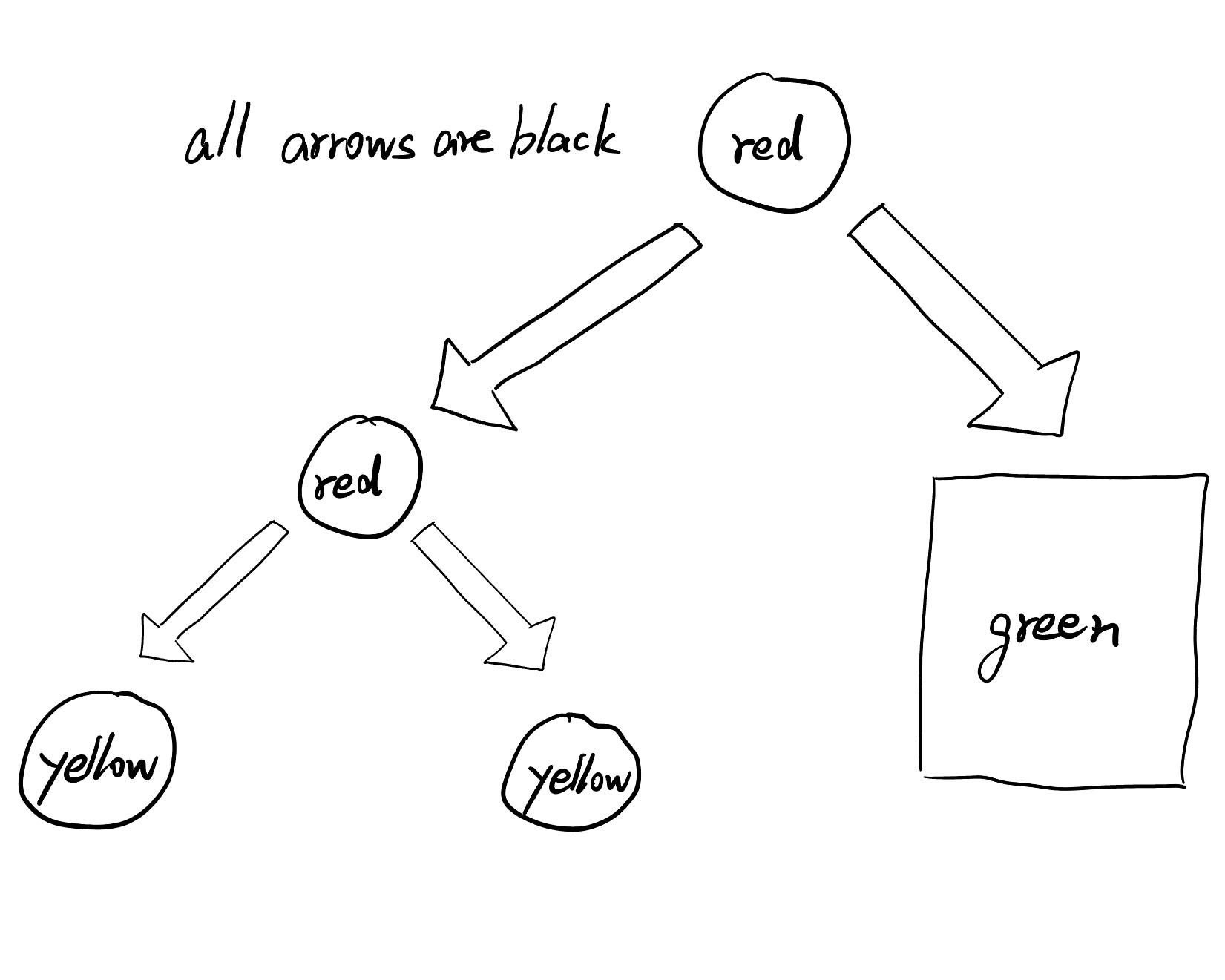} &
\includegraphics[width=\linewidth]{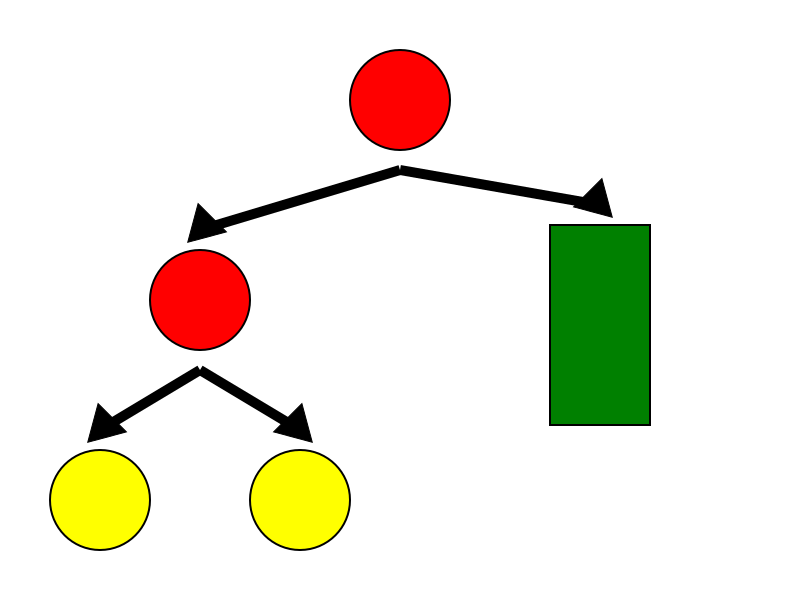} &
\includegraphics[width=\linewidth]{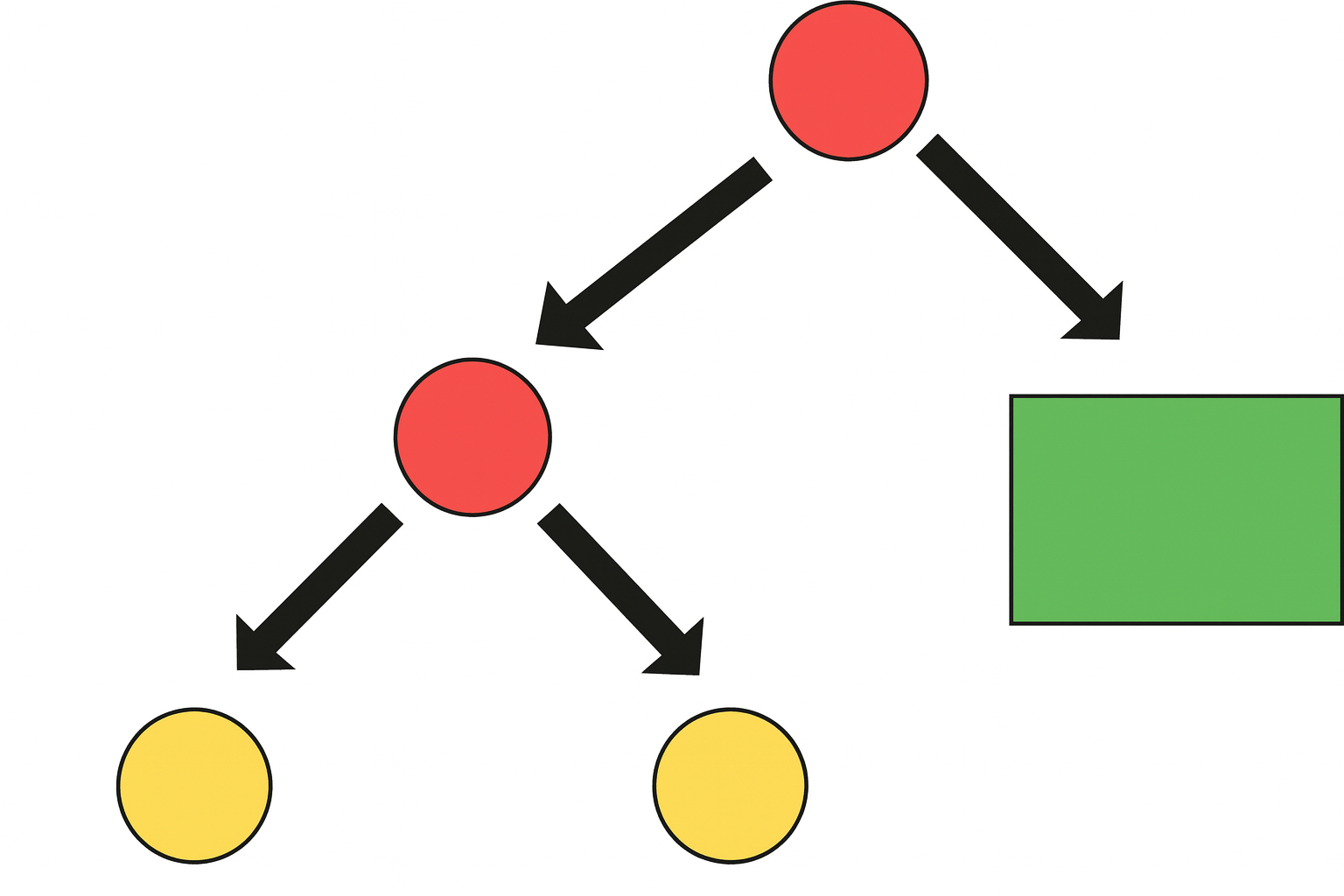} &
\includegraphics[width=\linewidth]{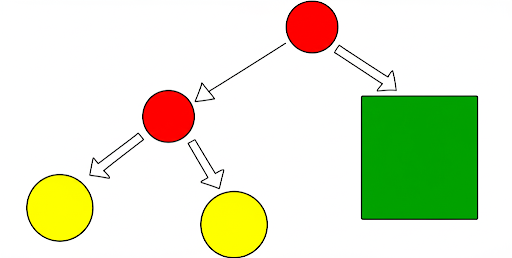} \\ \hline

\includegraphics[width=\linewidth]{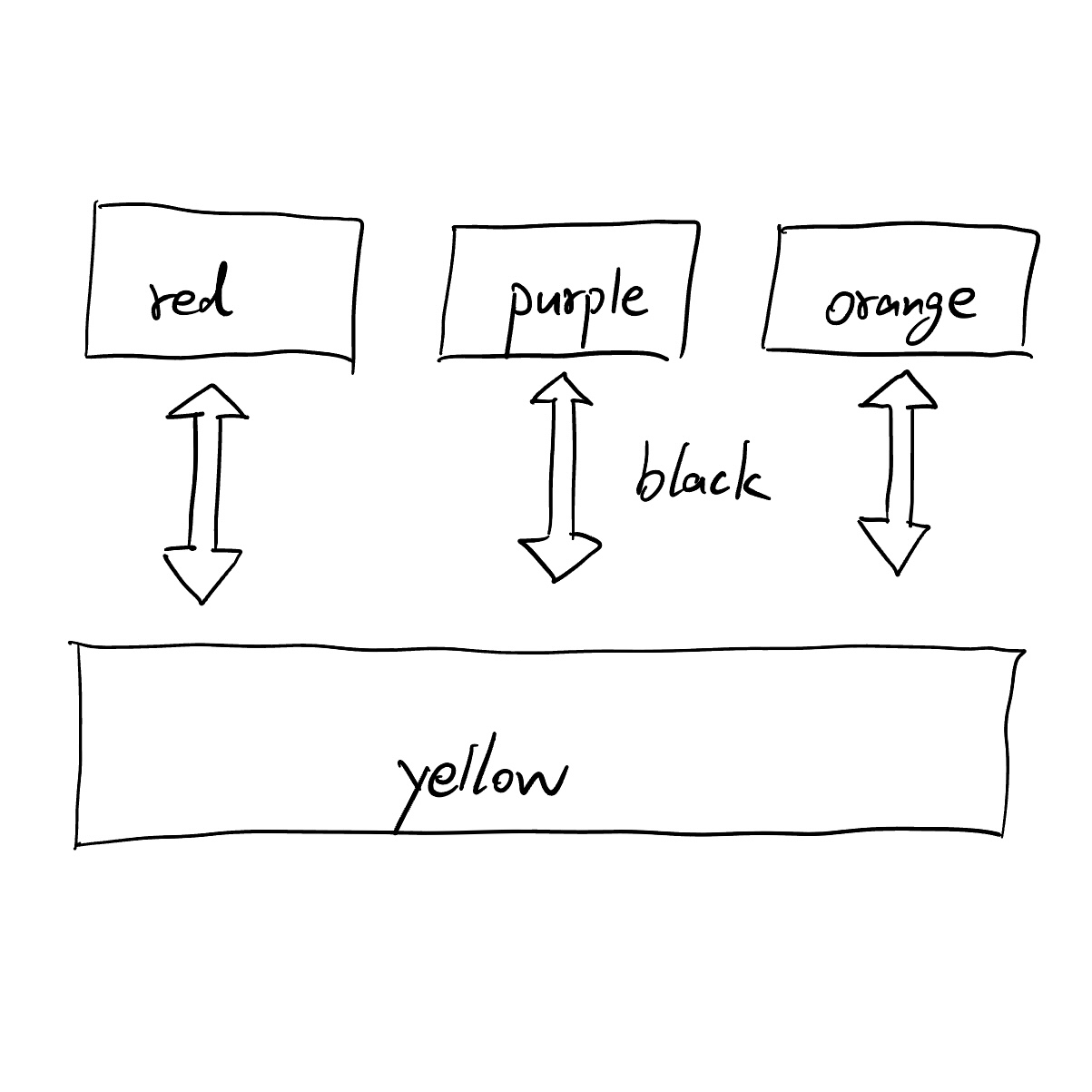} &
\includegraphics[width=\linewidth]{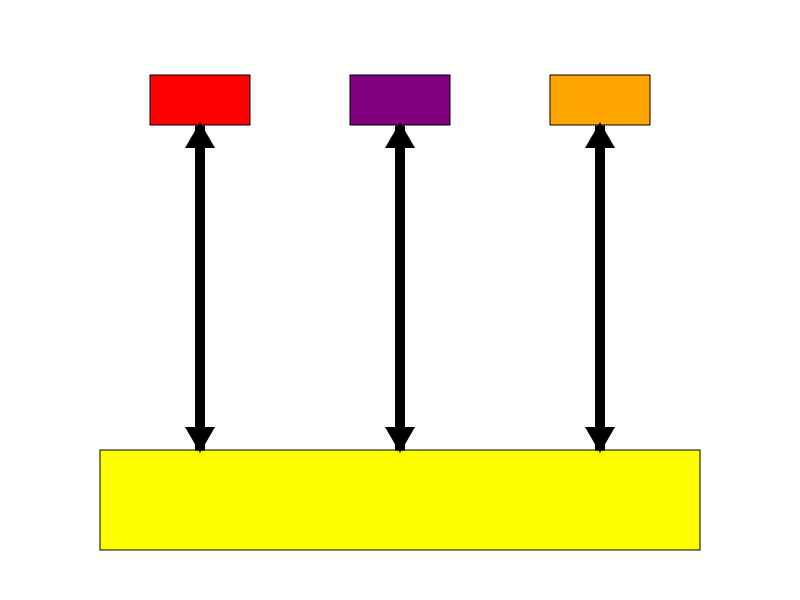} &
\includegraphics[width=\linewidth]{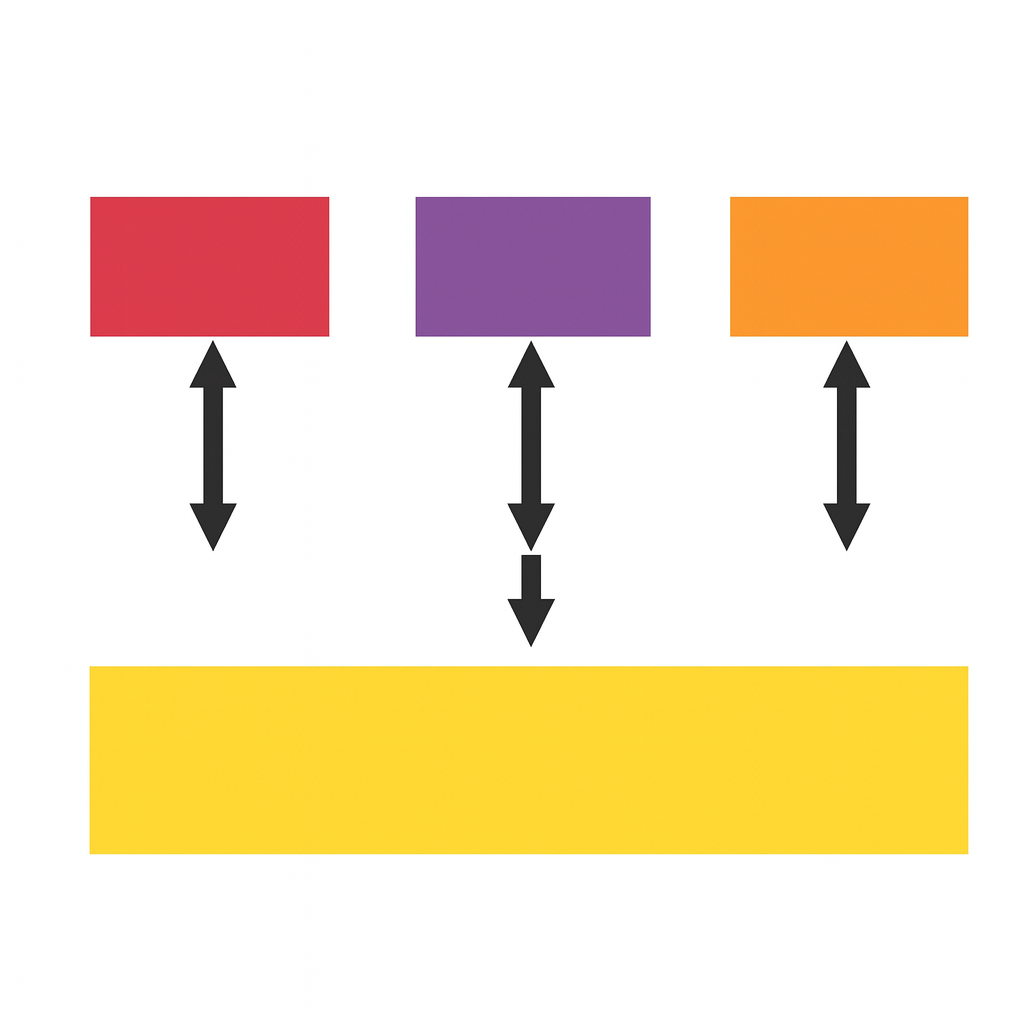} &
\includegraphics[width=\linewidth]{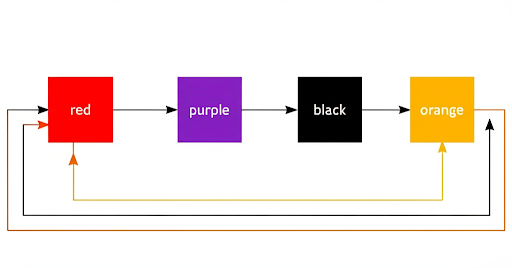} \\ \hline

\includegraphics[width=\linewidth]{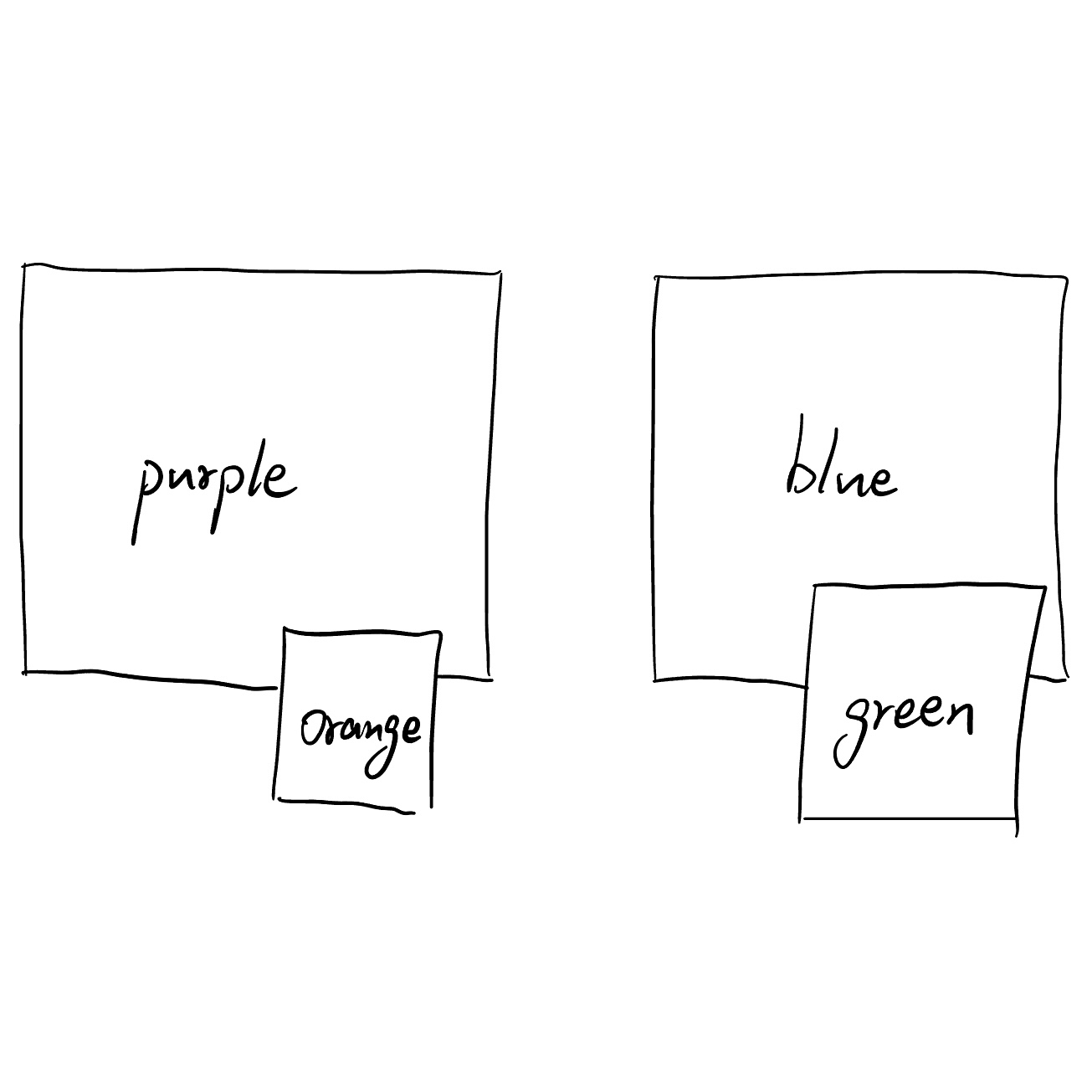} &
\includegraphics[width=\linewidth]{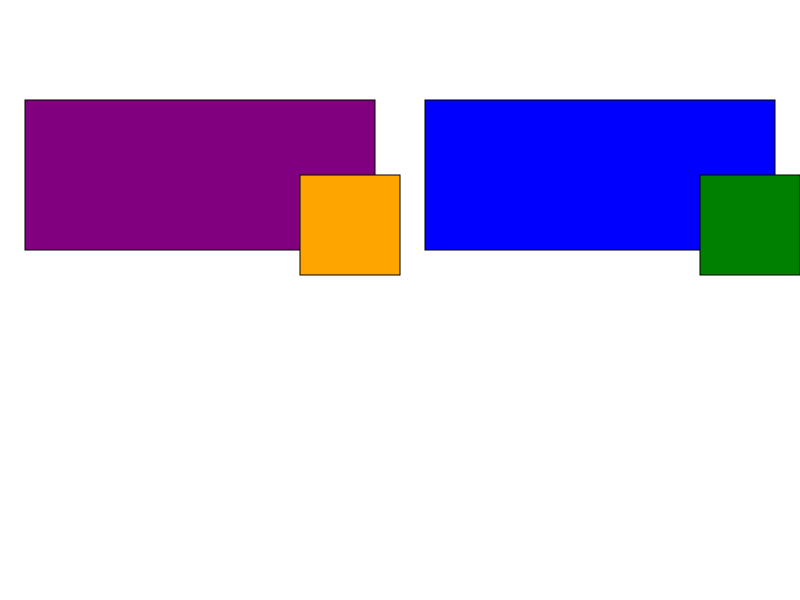} &
\includegraphics[width=\linewidth]{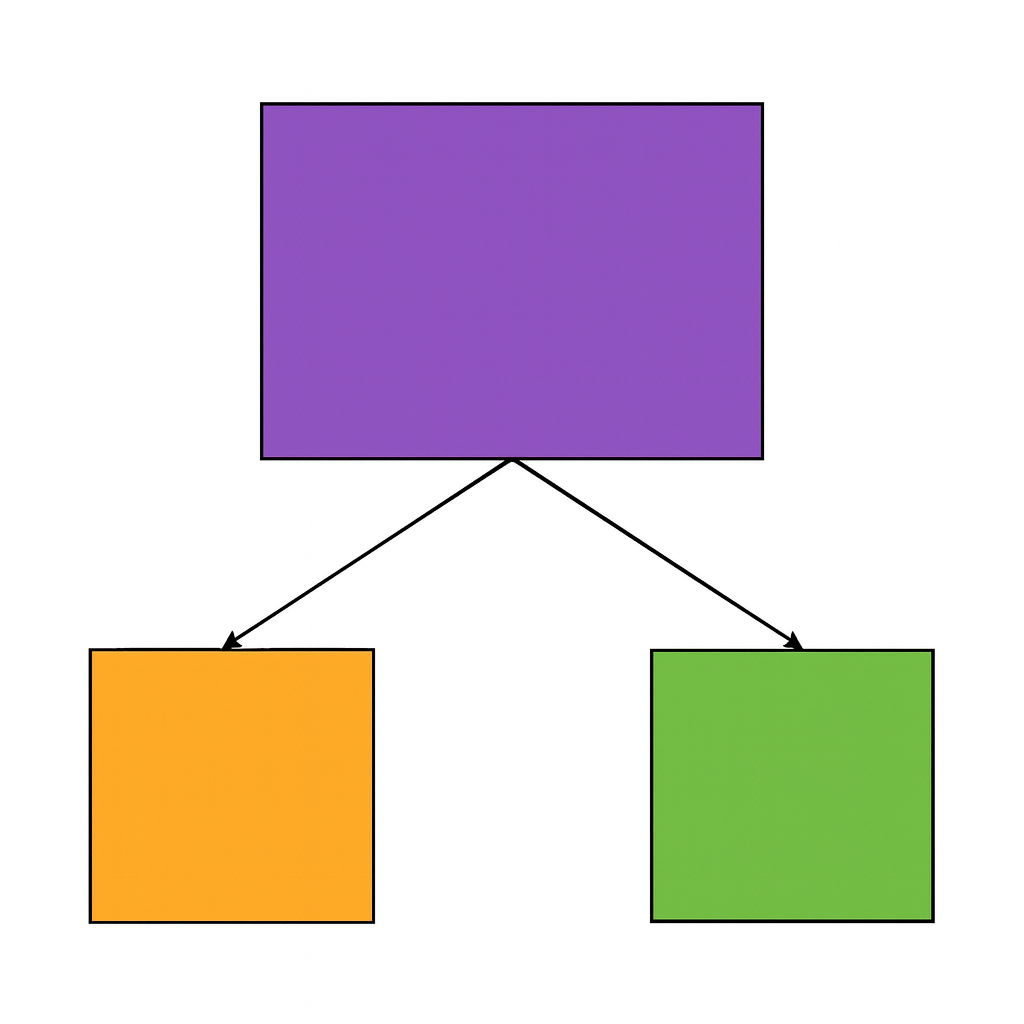} &
\includegraphics[width=\linewidth]{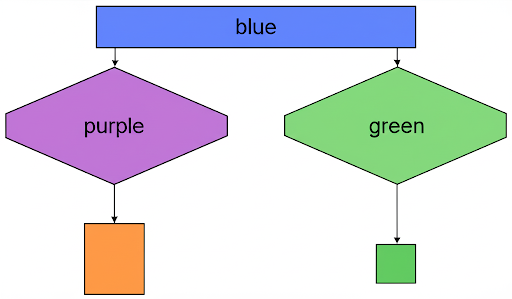} \\ \hline

\end{tabular}
\caption{Task 6-10}
\label{tab:comparison_2}
\end{table}

\newpage

\subsection{SVG Grammar}

\lstdefinestyle{promptbox}{
  basicstyle=\linespread{1.0}\ttfamily\small,
  backgroundcolor=\color{black!3},
  frame=single,
  rulecolor=\color{black!20},
  frameround=tttt,
  columns=fullflexible,
  keepspaces=true,
  showstringspaces=false,
  breaklines=true,
  tabsize=2
}

\noindent
\begin{minipage}{\linewidth}
\begin{lstlisting}[style=promptbox, caption={SVG grammar}]
Canvas
  size: {canvas_width} x {canvas_height}   # pixels (width x height)
  coords: origin (0,0) top-left; +x right; +y down
  positioning: (x,y) = center of shape

Types
  <type> := "circle" | "rectangle" | "ellipse" | "triangle"

Colors (lowercase only)
  <color> := "red"|"green"|"blue"|"yellow"|"purple"|"orange"|"black"|"white"|"none"
  # "none" disables fill or stroke respectively

Shape Object
  {
    "shape_type":   <type>,                           # required
    "x":            <num>=0,  "y": <num>=0,           # center position (px)
    "scale_x":      <num>=1,  "scale_y": <num>=1,     # width/height; for circle/ellipse: diameters
    "fill_color":   <color>="none",                   # fill color (use "none" for no fill)
    "stroke_color": <color>="black",                  # stroke color (use "none" for no stroke)
    "stroke_width": <num>=1,                          # stroke thickness (px)
    "rotation":     <num>=0                           # degrees, clockwise; triangle 0 degree points up
  }

Output (JSON)
  {
    "shapes": [ <Shape Object>, <Shape Object>, ... ]
  }
\end{lstlisting}
\label{svg_grammar}
\end{minipage}

\end{document}